\documentclass{article}

\usepackage{arxiv}

\usepackage[utf8]{inputenc} 
\usepackage[T1]{fontenc}    
\usepackage{hyperref}       
\usepackage{url}            
\usepackage{booktabs}       
\usepackage{amsfonts}       
\usepackage{nicefrac}       
\usepackage{microtype}      
\usepackage{lipsum}
\usepackage{graphicx}
\usepackage{natbib}
\setcitestyle{numbers}

\usepackage{balance} 
\usepackage{fancyvrb}
\usepackage{fvextra}
\usepackage[skins,breakable]{tcolorbox}

\newcommand{\frommak}[2][]{#1\frombody{blue}{mak}{#2}}

\definecolor{brown}{rgb}{.69,.37,.22}

\newcommand{\frombody}[3]{
	\noindent
	\textcolor{#1}{
		{$\bf [\!\![\!\![$}\underline{\scshape{#2}}
		{\scshape says:} 
		\textsl{#3}{$\bf ]\!\!]\!\!]$}}
	{}
}
\renewcommand{\frombody}[3]{}

\usepackage{algorithm}
\usepackage[noend]{algpseudocode}

\newcommand{\secref}[1]{\S\ref{#1}}
\newcommand{\appref}[1]{Appendix~\ref{#1}}
\usepackage{amsmath}

\usepackage{xspace}

\usepackage{enumitem}
\setitemize{noitemsep,topsep=0pt,parsep=0pt,partopsep=0pt,leftmargin=12pt}

\newcommand{\simplepar}[1]{\vspace{2mm} \noindent \textbf{#1}.}
\newcommand{\simpleparNoEnd}[1]{\vspace{2mm} \noindent \textbf{#1}}

\newcommand{\filename}[1]{\underline{\textsc{#1}}}  
\newcommand{\requestFile}{\filename{Request}\xspace}
\newcommand{\problemSpecFile}{\filename{ProblemSpec}\xspace}
\newcommand{\toolsSpecFile}{\filename{ToolsSpec}\xspace}
\newcommand{\toolsFile}{\filename{Tools}\xspace}
\newcommand{\solverpy}{\filename{Solver}\xspace}
\newcommand{\outputFile}{\filename{output}\xspace}
\newcommand{\errorFile}{\filename{error}\xspace}
\newcommand{\notesFile}{\filename{Notes}\xspace}
\newcommand{\answerFile}{\filename{Answer}\xspace}
\newcommand{\file}[0]{\filename{file}\xspace}

\newcommand{\systemname}{\textsc{ProcLLM}\xspace}

\newcommand{\mdpmaster}{\ensuremath{\mathcal{M}}\xspace}

\newcommand{\mdpstates}[1]{%
   \ifthenelse{ \equal{#1}{} }
      {\ensuremath{\mathcal{S}}\xspace}
      {\ensuremath{\mathcal{S}_{#1}}\xspace}}

\newcommand{\mdpstate}[1]{
   \ifthenelse{ \equal{#1}{} }
      {\ensuremath{s}\xspace}
      {\ensuremath{s_{#1}}\xspace}}
\newcommand{\mdpactionLetter}{\ensuremath{\mathcal{A}}}
\newcommand{\mdpactions}{\mdpactionLetter\xspace}

\newcommand{\mdpaction}{\ensuremath{a}\xspace}
\newcommand{\mdptransition}{\ensuremath{\mathcal{T}}\xspace}

\newcommand{\mdpreward}[1]{
    \ifthenelse{ \equal{#1}{} }
    {\ensuremath{r}\xspace}
    {\ensuremath{r_{#1}}\xspace}}

\newcommand{\verify}{\texttt{verify}\xspace}

\newcommand{\policy}{\ensuremath{\pi}\xspace}

\usepackage{yfonts} 

\newcommand{\tasks}[0]{\ensuremath{K}\xspace}
\newcommand{\task}[1]{\ensuremath{\kappa_{#1}}\xspace}
\newcommand{\taskname}[1]{{\footnotesize\texttt{#1}}\xspace}
\newcommand{\methodname}[1]{{\footnotesize\texttt{#1}}\xspace}
\newcommand{\methods}[0]{\ensuremath{\mathcal{M}}\xspace}
\newcommand{\method}[0]{\ensuremath{m}\xspace}

\newcommand{\Context}[0]{\ensuremath{C}\xspace}
\newcommand{\context}[1]{\ensuremath{C_{#1}}\xspace}
\newcommand{\contextAction}{\context{a}}
\newcommand{\contextVerify}{\context{v}}
\newcommand{\Horizon}[0]{\ensuremath{H}\xspace}
\newcommand{\actionLLM}[0]{\textnormal{action-LLM}\xspace}
\newcommand{\verifyLLM}[0]{\textnormal{verify-LLM}\xspace}
\newcommand{\averify}{\texttt{verify}\xspace}
\newcommand{\aread}{\texttt{read}\xspace}
\newcommand{\awrite}{\texttt{write}\xspace}
\newcommand{\aappend}{\texttt{append}\xspace}
\newcommand{\content}[0]{\texttt{content}\xspace}

\newcommand{\verified}{\texttt{verified}\xspace}
\newcommand{\feedback}{\texttt{feedback}\xspace}
\newcommand{\trace}{\texttt{trace}\xspace}

\newcommand{\exec}{\textit{Exec}\xspace}
\newcommand{\updateTask}[0]{UpdateTask\xspace}
\newcommand{\assemblyFunc}{\ensuremath{\textgoth{A}}\xspace}
\newcommand{\assemblyAction}[1]{\ensuremath{\assemblyFunc_{a}({#1})}\xspace}
\newcommand{\assemblyVerify}[1]{\ensuremath{\assemblyFunc_{v}({#1})}\xspace}

\algnewcommand{\LeftComment}[1]{\Statex \(\triangleright\) #1}

\newcommand{\noTN}{\texttt{No-TN}\xspace}
\newcommand{\humanTN}{\texttt{Human-TN}\xspace}
\newcommand{\llmTN}{\texttt{LLM-TN}\xspace}

\usepackage{bera}
\usepackage{listings}
\usepackage{xcolor}

\lstdefinelanguage{json}{
    basicstyle=\normalfont\ttfamily\scriptsize,
    commentstyle=\color{black}, 
    stringstyle=\color{gray}, 
    numbers=left,
    numberstyle=\scriptsize,
    xleftmargin=3.0ex,
    numbersep=2pt,
    showstringspaces=false,
    breaklines=true,
    frame=lines,
    string=[s]{"}{"},
    comment=[l]{:\ "},
    morecomment=[l]{:"},
    literate=
        *{0}{{{\color{numb}0}}}{1}
         {1}{{{\color{numb}1}}}{1}
         {2}{{{\color{numb}2}}}{1}
         {3}{{{\color{numb}3}}}{1}
         {4}{{{\color{numb}4}}}{1}
         {5}{{{\color{numb}5}}}{1}
         {6}{{{\color{numb}6}}}{1}
         {7}{{{\color{numb}7}}}{1}
         {8}{{{\color{numb}8}}}{1}
         {9}{{{\color{numb}9}}}{1}
}

\newtcolorbox{cvbox}[1][]{
    enhanced,
    after skip=8mm,
    title=#1,
    breakable = true,
    fonttitle=\sffamily\bfseries,
    coltitle=cyan,
    colbacktitle=gray!10,   
    titlerule= 0pt,         
    overlay={%
        \ifcase\tcbsegmentstate
        \or%
        \else%
        \fi%
    }
    colback = gray,         
    colframe = black!75     
    }

\title{
Procedural Knowledge Improves Agentic LLM Workflows
}

\author{Vincent Hsiao \\
NRC Postdoctoral Fellow, Naval Research Laboratory \\
Washington, DC \\
United States \\
\texttt{vincent.hsiao.ctr@us.navy.mil}
\And
Mark Roberts \\
Navy Center for Applied Research in AI, Naval Research Laboratory\\
Washington, DC\\
United States\\
\texttt{mark.c.roberts20.civ@us.navy.mil}
\And
Leslie Smith \\
Navy Center for Applied Research in AI, Naval Research Laboratory\\
Washington, DC\\
United States\\
\texttt{leslie.n.smith20.civ@us.navy.mil}}

\begin{document}

\maketitle 

\begin{abstract}
Large language models (LLMs) often struggle when performing agentic tasks without substantial tool support, prom-pt engineering, or fine tuning.
Despite research showing that domain-dependent, procedural knowledge can dramatically increase planning efficiency, little work evaluates its potential for improving LLM performance on agentic tasks that may require implicit planning. 
We formalize, implement, and evaluate an agentic LLM workflow that leverages procedural knowledge in the form of a hierarchical task network (HTN).
Empirical results of our implementation show that hand-coded HTNs can dramatically improve LLM performance on agentic tasks, and using HTNs can boost a 20b or 70b parameter LLM to outperform a much larger 120b parameter LLM baseline.  Furthermore, LLM-created HTNs improve overall performance, though less so.
The results suggest that leveraging expertise--from humans, documents, or LLMs--to curate procedural knowledge will become another important tool for improving LLM workflows.
\end{abstract}


\keywords{Large Language Models, Agentic Systems, Task Networks}


         
\newcommand{\BibTeX}{\rm B\kern-.05em{\sc i\kern-.025em b}\kern-.08em\TeX}



\pagestyle{fancy}
\fancyhead{}


\fancypagestyle{firststyle}
{
\cfoot{ \begin{small} \textbf{DISTRIBUTION STATEMENT A.} Approved for public release; distribution is unlimited. \end{small}}
}
\thispagestyle{firststyle}



\section{Introduction}
\label{sec:introduction}

Procedural knowledge generally describes a sequence for solving a problem and has played an important role in AI.
For example, spanning nearly 50 years of automated planning research, procedural knowledge has been effective for areas such as decomposing abstract tasks into concrete actions \cite{tate77.ijcai.generatingProjectNetworks,baccusYang94.aij.downwardRefinement},  control rules for search \cite{bacchusKabanza00.aij.usingTemporalLogics}, learning macro operators \cite{colesSmith07.jair.marvin}, plan recognition \cite{geibGoldman09.aij.probabilisticPlanRecognition}, knowledge-based planning \cite{langleyShrobe21.acs.hierarchicalProblemNetworks}, and planning and acting \cite{ghallabNauTraverso25.book.apl}, to name only a few.
Procedural knowledge also plays a central role in cognitive systems  \cite{lairdLebiereRosenbloom17.aimag.standardModelOfMind}, where a specific module in the system manages and applies such  knowledge to problem solving.
Recently, procedural knowledge is used as a kind of meta-strategy to assist in solving complex problems (e.g., \cite{belcak2025universal}). 

While LLMs have shown impressive capability, they often perform poorly on planning problems \cite{kambhampati2024can}, and, as we will show later, LLM performance generally degrades as task complexity increases. 
In complex agentic tasks, successful task completion often hinges on an implicit plan that specifies what actions to take and in what sequence to take them, a capability that current LLMs often fail to consistently complete correctly.
Part of the gap is that LLMs are mostly constructed to predict the next token, while planning requires managing details that span different temporal horizons and different levels of abstraction. 
LLMs lack a clearly understood way to track the change that results from taking action in the world.


In contrast, this paper examines whether providing \emph{explicit} procedural knowledge improves LLM performance on agentic tasks.
%
To provide an intuitive example (formalized in \secref{sec:formalism-new}), consider the task of booking a travel itinerary.
Experienced travelers follow a sequence such as: book flights, book hotel, and reserve attractions or restaurants. 
The sequence might vary by circumstance, but its presence makes the problem much easier to solve.
The sequence manages dependencies (e.g., confirm flights before booking the hotel) and each step narrows the details considered in subsequent steps.
But using an abstract sequence requires a process that: 
1) tracks progress on the overall procedure, 
2) maintains state changes across steps, 
3) takes actions to complete each step, 
4) verifies each step is finished before proceeding.

When solving a travel problem like this, LLMs lack the capability to track these layers all at once, leading to inefficient or unsuccessful workflows. 
For example, an LLM might book a restaurant reservation at the destination before verifying flight timing, only to backtrack to address such foreseeable conflicts. 
By treating complex tasks as a series of independent decisions rather than attending to interconnected dependencies, LLMs struggle to leverage procedural (i.e., hierarchical, goal-oriented) knowledge.

In this paper, we address two questions: \emph{Can we embed procedural knowledge in an LLM workflow?} and \emph{Will procedural knowledge improve LLM performance?}
We will show that the answer to both questions is \emph{Yes!}~~To that end:
\begin{itemize}
    \item We formalize the problem as an MDP and integrate procedural knowledge using Hierarchical Task Networks (HTNs), where abstract tasks decompose into totally ordered subtasks.
    An HTN 
    ensures logical consistency and resource availability.

    \item We describe and implement a hybrid LLM-HTN system, \systemname{}, that uses HTNs in an agentic LLM workflow.  
    Importantly, this workflow is compatible with any LLM. 

    \item We evaluate the runtime, success, and number of cycles of \systemname{} on four benchmark problems, two from the literature and two synthetic problems. \textbf{\emph{For the problems and HTNs we studied, HTN knowledge always improves LLM performance, often significantly.}} \frommak{triple check this claim; adjust as needed}
    
    \item We demonstrate that smaller LLMs benefit more from this workflow and in some cases exceed the performance of much larger LLMs without HTN knowledge.  

    \item Finally, we show that LLM-created HTNs improve performance, though the results are mixed compared to hand-coded HTNs.
    
\end{itemize}

\noindent
Overall, these findings suggest that LLM workflows can benefit considerably from procedural knowledge, not only in overall performance but also in a reduced model size and improved response time for the same level of performance.  
Although our implementation uses an Agentic LLM workflow with HTN knowledge, the process is straightforward and results may eventually generalize to other LLM workflows, so we close with some possible future directions.

\section{Related Work}

\simplepar{Agentic LLMs}
We define agentic LLM as a large language model framework that can  act, use tools, and plan to autonomously accomplish tasks.
Many problems cannot be solved in a single forward pass of an LLM, motivating workflow techniques such as prompt chaining \cite{wu2022promptchainer}, where the output of one prompt becomes the input to another.
Early community systems like AutoGPT \cite{yang2023auto} showcased the potential of chaining and memory-augmented reasoning for open-ended tasks, while frameworks such as LangChain provided modular infrastructure for building agentic pipelines around LLMs

One of the first academic approaches in this area was the ReAct framework \cite{yao2022react}, where agent reasoning choices are interleaved with action selection in order to interact with external environments. Another notable contribution was ExpeL \cite{zhao2024expel}, where an agent that learns from experiences and natural language to make informed decisions without requiring parametric updates. 

Although there are many variations of this framework, for the purpose of this paper, we will use a framework similar to the one used by Reflexion (Fig. 2 in \cite{shinn2023reflexion}). The Reflexion agent employs an LLM (denoted as the agent) and an environment that processes actions across multiple iterations. At each iteration, the agent receives an observation from the environment. This observation is combined with a prompt given to the LLM to output some action. The environment processes this action and then returns the observation for the next iteration. 

Perhaps the closest to this paper is the concurrent work of Belcak and Molchanov \cite{belcak2025universal}. In their universal deep research (UDR) framework, a strategy compiler is used to take a research strategy from the user and compile it into a list of steps and corresponding outputs that a deep research tool needs to execute step by step. The structure is enforced by incorporating code comments and yield directives that specify the goal of each step and the intended outputs of the step. Our approach is more general, and we can hypothesize that there exists a task network that when applied to our system would replicate the behavior of UDR. 

\simplepar{LLMs and Tool-use}
To supplement capabilities in domains where LLMs have poor performance, it is common to allow LLMs to interface with external tools. These tools enhance the LLMs with capabilities that they might otherwise not have \cite{wang2024tools} and there are numerous papers covering various aspects of this field of research \cite{huang2024understanding, li2024survey}.  Early demonstrations included Toolformer \cite{schick2023toolformer}, which fine-tuned LLMs to invoke APIs from demonstrations, and MRKL \cite{karpas2022mrkl}, which integrated symbolic reasoning with tool-use. 

Unlike past work, our system allows the LLM agent to write arbitrary code to call arbitrary APIs. This is different from works where the external tools are presented in a standardized in-context format to an LLM for usage (e.g., MCP servers). Allowing the LLM to write arbitrary code also gives more flexibility in problem solving (e.g., coding allows the LLM to embed tool calls in algorithms), which standardized tool formats do not permit. 

\simplepar{Multi-agent Systems}
The extension from single-agent to multi-agent frameworks has become an important line of research. In AgentOrchestra \cite{zhang2025agentorchestra}, an orchestrator agent creates a plan that is then executed by specialized sub-agents. Unlike our explicit task network representation, AgentOrchestra uses a simple list structure. SagaLLM \cite{chang2025sagallm} proposes a transactional system to handle multi-step planning in multi-agent LLM systems. Other works, such as CAMEL \cite{li2023camel}, demonstrated emergent coordination between specialized LLM agents by role-playing tasks and negotiating via natural language.

\simplepar{Hybrid Systems}
Various hybrid systems have been developed for application to specific domains. For example, \cite{huang2025towards} provides a survey of LLM-based agentic recommenders that incorporate subsystems similar to planning and acting modules. AutoConcierge \cite{zeng2024automated} employs LLMs to convert user requests into formalized representations to be used by a reasoner. This is also similar to the approach from \cite{hao2024large}, where user requests and API calls are translated into constraints for an SMT solver. Both approaches use LLMs to convert user data into formalized problem descriptions, which are then solved to provide a solution. This is different from our approach, which instead uses a formalized representation to assist the LLM in the problem-solving process.

Other than agentic LLM systems, there are many LLM-modulo systems that combine LLMs with external verifiers and or solvers \cite{kambhampati2024position}. ISR-LLM \cite{zhou2024isr} employs an LLM to translate planning problems in natural language into PDDL and then iteratively calls the external planner to verify LLM-generated plans. 
Beyond agentic LLM systems, many hybrid approaches combine LLMs with external verifiers or solvers \cite{kambhampati2024position}. ISR-LLM \cite{zhou2024isr}, for instance, employs an LLM to translate planning problems in natural language into PDDL and then iteratively calls an external planner to verify and refine plans. Similarly, program synthesis approaches \cite{chen2021evaluating} use LLMs to generate candidate code that is executed and verified externally. These hybrid formulations highlight a broader paradigm: using LLMs as flexible natural-language interfaces while delegating correctness and reliability to symbolic or algorithmic components.

\simplepar{LLMs and Planning}
In challenging tasks, an agent may need to generate a plan before completing the task. Many researchers have investigated how LLMs can be used effectively for planning \cite{song2023llm} \cite{silver2024generalized} \cite{kambhampati2024position}. There have been impressive gains in this area of research; for example, a recent notable example is by \citet{vermaEtal25.arxiv.teachingLLMs} which claims up to 94\% planning accuracy using a fine-tuning approach paired with chain-of-thought reasoning.  Despite these gains, planning remains a challenging task \cite{shojaee2025illusion} \cite{valmeekam2024planning} with many results showing that LLMs still have many weaknesses when applied to these problems.  Furthermore, many approaches still require considerable scaffolding.

\frommak{there was a quote before universal -- was it an actual quote?  Or was this a typo?}
A consistent claim is that the effectiveness of LLM planning is reduced as task complexity increases. \citet{kambhampati2024can} argues that LLMs engage in universal approximate retrieval, where they rely on pattern matching rather than systematic reasoning. This suffices for generating immediate solutions or short inference chains, but can run into issues when task complexity increases, which can be seen in ACPBench \cite{kokel2025acpbench} where the task of generating just the next action has a much higher success rate than the more complex tasks. This suggests that current LLMs are most effective at solving short tasks that do not need multi-step planning. A natural solution would be as follows: When the provided task is too complex for an LLM agent, decompose the task into subtasks for which the LLM is effective.

Unlike other works in this area, our focus is not directly on the ability of an LLM to produce a plan (i.e., a sequence of steps solving the planning task). Rather, in many tasks that agentic LLMs are applied to solve, a sort of meta-planning is a requisite process for solving complex problem instances. By decomposing a task into smaller subtasks, the meta-planning required will be less complex and the agentic system will be less impacted by weaknesses of current LLMs when encountering complex tasks. 

\section{Preliminaries}
\label{sec:formalism-new}

We represent procedural knowledge using a  Hierarchical Task Network (HTN), described next.  
We then formalize the problem as a Markov Decision Process (MDP) where the policy that solves the MDP is an LLM. 
Finally, we analyze why we expect \systemname{} to perform well.

\begin{figure}
     \begin{center}
     \includegraphics[width=0.7\columnwidth]{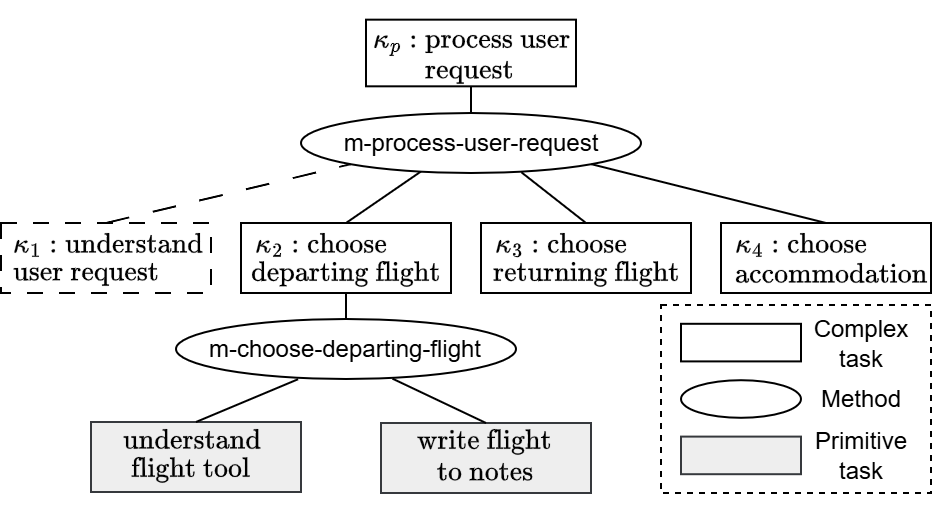}
     \end{center}
\begin{lstlisting}[language=json]
{"m-travel-process-user-request": { 
    "task": "process user request",
    "subtasks": {"1": "understand user request",
                 "2": "choose a departing flight",
                 "3": "choose a returning flight",
                 "4": "choose an accommodation"   },
    "verify": "answer.txt contains a solution in the correct format that follows the problem specification",
    "verify_files": {"file1": "problem_specification.txt",
                     "file2": "answer.txt"  }},
"m-choose-departing-flight": {
    "task": "choose a departing flight",
    "subtasks": {"1": "understand flights tool",
                 "2": "write flight to notes" },
    "verify": "notes contains details about a flight number and cost for a flight from origin to destination on departure date consistent with request preferences",
    "verify_files": {"file1": "files/notes.txt",
                     "file2": "files/problem_specification.txt",
                     "file3": "files/request.txt"  } },
"m-understand-flights-tool": {
    "name": "understand flights tool",
    "effect": "notes contains details about flights tool methods inputs and outputs",
    "effect_files": {"file1": "files/notes.txt",
                     "file2": "files/tools_specification.txt" } } }



\end{lstlisting}
     \caption{Top: An in-progress HTN decomposition tree for the Travel Planner problem of  \secref{sec:introduction}.  White boxes indicate abstract tasks, round ovals indicate methods that decompose tasks, and gray boxes indicate primitive tasks for the LLM to execute. The dashed box for \task{1} indicates it has passed \verify.
     Bottom: Notional methods for the same problem.
     }
     \label{fig:travelhtn}
 \end{figure}

\simplepar{Hierarchical Task Networks (HTNs)}
Because our focus is on how procedural knowledge improves LLM performance, our description of HTNs sacrifices exact formality for a more notional understanding.  A full formal treatment can be found in \cite{ghallabNauTraverso25.book.apl}.
Let \tasks{} be a set of tasks; a task $\task{} \in \tasks$ is a labeled name of something the system needs to do. 
Figure~\ref{fig:travelhtn} (top) shows an HTN decomposition tree where \tasks{} relates to the travel problem from \secref{sec:introduction}. 
HTNs decompose complex tasks (white boxes) into  primitive tasks (gray boxes) using methods (ovals).
\emph{Complex tasks} describe abstract tasks that need to be decomposed and
\emph{primitive tasks} describe concrete executable tasks.
Figure~\ref{fig:travelhtn} shows two methods. 
Method \methodname{m-process-user-request} decomposes \task{p} into $\langle \task{1}, .., \task{4} \rangle$, where the \task{1} (dashed box) has passed the \verify check (described below).  
Method \taskname{choose departing flight} decomposes \task{2} into primitive tasks: $\langle \taskname{understand flight tool}, \taskname{write flight to notes} \rangle$.

For a set of methods \methods{}, a method $\method \in \methods$ is a tuple $(head, task, pre, subtasks)$, where the $head$ is the name and any paramaters of the method,  $task$  matches the taskname the method can decompose, $pre$ is a (possibly empty) set of preconditions for applying the method in a state, and $subtasks$ is a totally ordered sequence of tasks $\langle \task{1}, \task{2}, .., \task{n} \rangle$.  
A method that matches a task is \emph{relevant}, and HTN planners often select the first relevant method in \methods{} (As described in \secref{sec:system-description}, \systemname{}'s Algorithm~\ref{alg:procllm}, Line~\ref{line:find-relevant} uses FindFirstRelevantMethod to select methods).

\simplepar{Markov Decision Process (MDP) Model }
\frommak{I think this formalism can work but it may make sense to eventually shift this to the model given in: \url{https://www.ijcai.org/proceedings/2022/0770.pdf}. that might also help with a shift to HGNs.}
We formalize agentic tasks as a finite-horizon deterministic Markov Decision Process (MDP), defined by the tuple $\mdpmaster = (\mdpstates{}, \mdpactions{}, \mdptransition, \mdpreward{}, \Horizon)$, where 
$\mdpstates{}$ is the set of states with a subset of absorbing terminal states, $\mdpstates{\text{term}} \subset \mdpstates{}$,
\mdpactions{} is set of actions,
$\mdptransition: \mdpstates{} \times \mdpactions{} \rightarrow \mdpstates{}$ is a deterministic state transition function, where $s_{i+1} = \mdptransition(\mdpstate{i}, \mdpaction_i)$,
$\mdpreward{}: \mdpstates{} \times \mdpactions{} \rightarrow \mathbb{R}$ is a reward function,
and $\Horizon$ is a finite horizon. An episode terminates either by reaching a terminal state or by exceeding the horizon (timeout).

For our domains, we define a function $\verify: \mathcal{S} \rightarrow \{0, 1\}$, that identifies terminal states (i.e., $\verify(\mdpstate{}) = 1$ if $\mdpstate{} \in \mdpstates{\text{term}}$ and $0$ otherwise). The reward function is then:
\[
\mdpreward{}(\mdpstate, \mdpaction) = 
\begin{cases} 
    \mdpreward{\text{success}} & \text{if } \verify(\mdptransition(\mdpstate, \mdpaction)) = 1 \quad \text{(a terminal state)} \\
    \mdpreward{\text{step}} & \text{otherwise}
\end{cases}
\]
where $\mdpreward{\text{success}} = 1$ is a positive reward for task completion and $\mdpreward{\text{step}}$ is a negative reward for non-terminal steps (e.g., $\mdpreward{\text{step}} = -0.1$).
Primitive tasks from the HTN are linked to \mdpstates{term}, which can be verified using \verify{}.

\frommak{I corrected one $a_t$ into $a_i$}
An agent's behavior is described by a stochastic policy $\policy{}: \mdpstates{} \times \mdpactions{} \rightarrow [0,1]$, where $\policy{}(\mdpstate{}, \mdpaction) = P(\mdpaction_i = \mdpaction \mid \mdpstate{i} = \mdpstate{})$ is the probability of selecting $\mdpaction$ in state $\mdpstate{}$. A solution to $\mdpmaster$ is a policy $\policy{}$ that maximizes the expected reward (in this case a policy that reaches a terminal state in the fewest steps).




\simplepar{Initial Analysis}
In this work, we investigate the use of an agentic LLM system as the policy $\policy{}$ that operates within this MDP framework. We show that divide-and-conquer mechanisms such as task decomposition can be used to constrain $\policy{}$ to work on smaller problems, increasing the expected reward obtained by our agentic LLM. 

The general MDP model allows us to hypothesize about why we expect this approach to work. We assume that there exists a task-aware assembly function $\assemblyFunc : \mdpstates{} \times \tasks{} \rightarrow \context{}$ and let $\context{i}$ be the context produced from state $\mdpstate{i}$ used as input to the LLM at each iteration.

\simplepar{Task Success Rate}
A task succeeds if a terminal state $ \mdpstate{} \in \mdpstates{\text{term}} $ is reached within $ H $ steps. Let $(\mdpaction_0, \dots, \mdpaction_{\Horizon-1})$ be a sequence of actions that arrives at a terminal state from an initial state $\mdpstate{0}$. Assuming deterministic transitions, the probability of taking these actions depends solely on the LLM's parameterization $ \theta $, which governs action selection:
$$ P_\theta(\mdpaction_0, \mdpaction_1, \dots, \mdpaction_{\Horizon-1}) = \prod_{i=0}^{\Horizon-1} P_\theta(\mdpaction_i | \context{i})
$$
\frommak{I corrected two $a_t$ into $a_i$}
Here, $ P_\theta(\mdpaction_i | \context{i}) $ represents the LLM's conditional probability of selecting action $ \mdpaction_i $ given context $ \context{i} $. This factorization arises because each action depends only on the current context (Markov property).

\simplepar{Task Decomposition}
A task decomposition splits a problem into a sequence of smaller subtasks, $(\task{0}, \dots, \task{n-1})$. Let $P_\theta(\task{j})$ be the probability of successfully completing subtask $j$. Decomposition improves task success if:
$$
P_\theta(\text{full task}) < \prod_{j=0}^{n-1} P_\theta(\task{j})
$$
We hypothesize that we can improve the right side of this equation through contextual scaffolding. We assume that the LLM's ability to complete a subtask $ P_\theta(\task{j}) $ increases by reducing context complexity. When the agent works on a specific subtask $\task{j}$, the assembly function $\assemblyFunc(s_i, \task{j})$ generates a focused context $\context{i,j}'$. By constraining the problem space through reducing context complexity, the conditional probability of selecting a correct action $\mdpaction_i$ should increase:
$$
P_\theta(\mdpaction_i | \context{i,j}') > P_\theta(\mdpaction_i | \context{i})
$$
Because the probability of success for a subtask $\task{j}$, $P_\theta(\task{j})$, is the product of the individual $P_\theta(\mdpaction_i | \context{i,j}')$, the task success rate $P_\theta(\task{j})$ will also increase.


\section{\systemname{}: Embedding Procedural knowledge into an agentic LLM }
\label{sec:system-description}
\begin{figure}
     \begin{center}
     \includegraphics[width=0.8\textwidth]{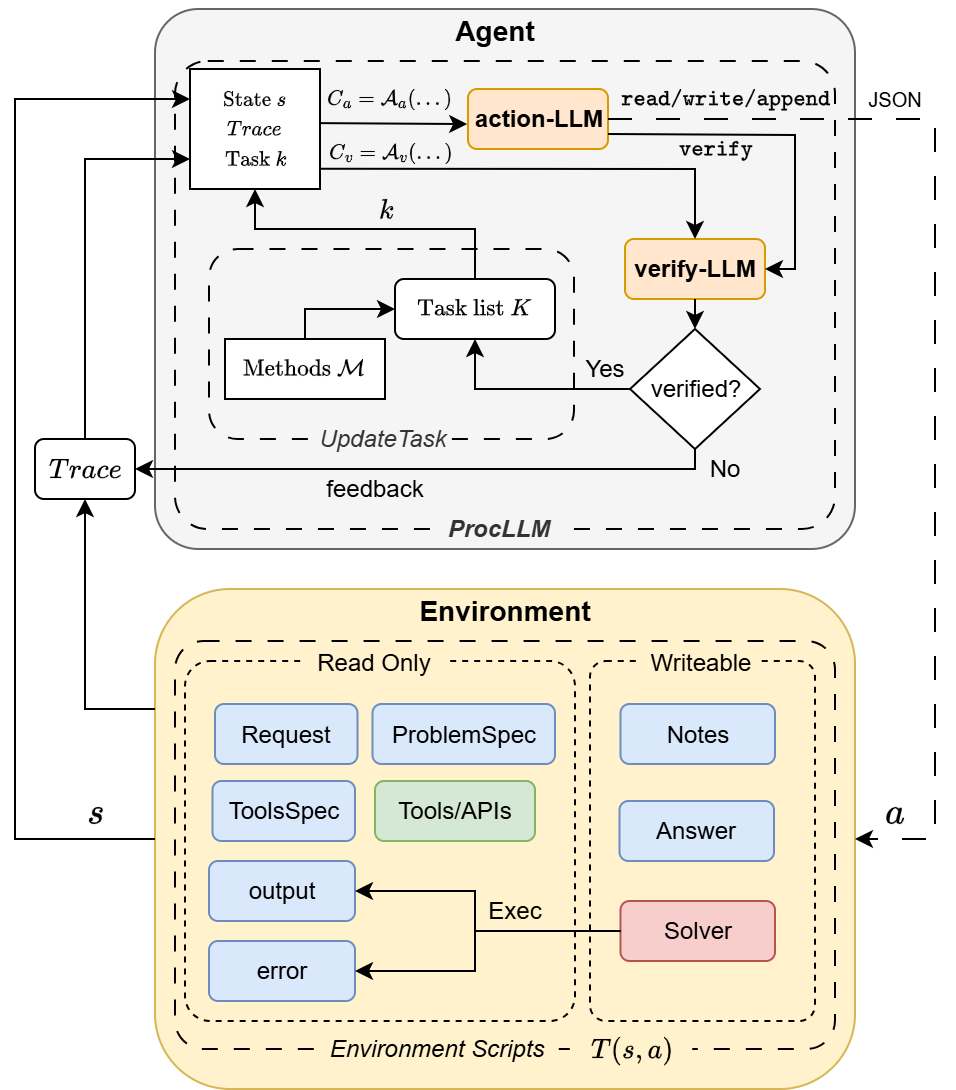}
     \end{center}
     \caption{System overview, labeled with components of the MDP. Blue boxes denote text files, orange boxes denote LLMs, red boxes denote python files, and green boxes denote API files (scripts/databases/etc.). 
     }
     \label{fig:system-workflow}
 \end{figure}

\systemname{} implements a planning and acting system \cite{ghallabNauTraverso25.book.apl} in which an agent iterates between HTN planning, where it updates the current (sub)task based on progress, and action execution, where it executes primitive tasks (i.e., actions) and verifies that they are completed.  
Relating \systemname{} to the formalism, we next describe the components of this system, as shown in Figure~\ref{fig:system-workflow}.

\simplepar{Environment and states \mdpstates{}} 
For the benchmarks we study in this paper, the environment corresponds to a file system with text files and python code files (Figure~\ref{fig:system-workflow}, bottom in yellow). 
The agent is provided with the following text (.txt) files or python (.py) code, where the first column \texttt{rwa} indicates allowed read/write/append actions, respectively.  
\begin{itemize}[noitemsep,topsep=0pt,parsep=0pt,partopsep=0pt,leftmargin=20pt]
    \item[r] 
    \problemSpecFile outlines the domain-specific rules of the problem to solve and format of \answerFile, which is the same for all request instances. (See~\secref{app:domain-spec} for full details.)
    \item[r] 
    \requestFile contains problem instance details to be solved such as the initial configuration of the problem and the goal configuration.    (See~\secref{app:domain-spec} for full details.)
    \item[r] 
    \toolsSpecFile is a human readable description of \toolsFile, 
    which are APIs that include source code, informational text files, or structured text (e.g.,  csv, JSON).
    \item[rwa] 
    \notesFile starts empty and will contain the agent's progress or details about the task or task history.
    \item[rwa] 
    \solverpy is only provided for some problems; the environment executes it, writing to \outputFile and \errorFile.
    \item[rwa] 
    \answerFile starts empty and will contain the agent's solution to the overall task. 
\end{itemize}

The content of the files as well as other details are assembled into context 
$\Context = \assemblyFunc(e_1, e_2, ..., e_k)$ containing one or more elements $e_i$.  
For \systemname{} this includes the action context $\contextAction = \assemblyAction{\mdpstate, \trace, \task{})}$, and the verify context $\contextVerify = \assemblyVerify{\mdpstate{}, \trace, \task{}}$.

\begin{algorithm}
    \caption{A general procedure for \systemname{}. 
    \newline
    \textbf{Input:} Task Sequence \tasks, Methods \methods, horizon \Horizon
    }
    \label{alg:procllm}
\begin{small}
    
    \begin{algorithmic}[1]
        \Procedure{\systemname{}}{\tasks, \methods, \Horizon}
            \State \updateTask(\tasks, \methods) \label{line:updateTask}
            \For{$i$ in $1, ..., H$}  \label{line:until-horizon}
                \If{$\tasks==\emptyset$}  \label{line:empty-task}
                    break
                \EndIf
                \State $\task{} \gets head(\tasks)$  \label{line:current-task}
                \State $\contextAction \gets \assemblyAction{\mdpstate, \trace, \task{}}$ \label{line:action-context}
                \State $\mdpaction(params) \gets \actionLLM(\contextAction)$ \label{line:action-llm}
                \State $\trace \gets $ `'
                \If{$\mdpaction = \averify$}  \label{line:verify-start}
                    \State $\contextVerify \gets \assemblyVerify{\mdpstate, \task{}}$  \label{line:verify-context}
                    \State \verified, \feedback$~\gets \verifyLLM(\contextVerify)$ \label{line:verify-llm}
                    \State $\trace \gets \feedback$
                    \If{\verified}
                        \State $pop(\tasks)$ \Comment{remove first task \task{} from \tasks}
                        \State \updateTask(\tasks, \methods)  \label{line:verify-end}
                    \EndIf

                \Else  \> $\triangleright$ apply r/w/a to env't; (performs $\mdpstate{}' \gets \mdptransition(\mdpstate, \mdpaction))$ \label{line:external-start}
                    \State $\trace \gets$ file \Comment{write file to the trace }
                    \If{$\mdpaction = \awrite($file, content$)$}
                        file.clear$()$
                    \EndIf
                    \State file $\gets$ file + content
                    \If{file$~= \solverpy$}
                        \State out, err $\gets \exec(\solverpy)$
                        \State $\trace \gets \trace~+ $ out $ + $ err  \label{line:external-end}
                    \EndIf
                \EndIf
                
          \EndFor
        \EndProcedure
        \vspace{1mm}
        \Procedure{\updateTask}{\tasks, \methods} \label{line:update-tasks-start}
            \State $\method{} \gets $ FindFirstRelevantMethod$(head(\tasks{}), \methods)$ \label{line:find-relevant}
            \If{\method{} }
                 \State $push(\method{}.subtasks, \tasks{})$ \Comment{Prepend $\method{}.subtasks$} 
                 \State \updateTask(\tasks{}, \methods{}) \Comment{Decompose further} \label{line:update-tasks-end} 
            \EndIf
        \EndProcedure
    \end{algorithmic}
\end{small}
    \end{algorithm}
\begin{algorithm}[t]
\end{algorithm}

\simplepar{Actions \mdpactions{} and transition \mdptransition} 
The agent can take one of four actions: \averify, \aread, \awrite, or \aappend.  

The internal action \averify calls a process that checks if the current task has been satisfied, returning a boolean \verified if the task is completed as well as \feedback which is added to the \trace.
The \verify step is domain-specific and uses the \verifyLLM.  
(A detailed example of the \verify prompt is provided in \appref{app:verification}.)

The external actions modify the writeable files, which are subsequently assembled into the context.  These include:
\begin{tabular}{ll}
 \quad  & \aread(\file):  copies \file into \notesFile \\
   & \awrite(\file, \content):  overwrites \file with \content \\
   & \aappend(\file, \content): appends \content to \file \\
\end{tabular}\\
These external actions must be in a JSON format for the environment to execute (for a detailed example, see \appref{app:action-examples}).
The environment implements a transition function as a set of scripts that calls $\exec(\solverpy)$ and otherwise modifies files as indicated in the actions.

\simplepar{Tasks \tasks{} and Methods \methods} 
For \systemname{}, the tasks and methods are domain-dependent. 
Most domains will have the task \taskname{process user request}, which is the top level task that the agent will start with. 
Figure~\ref{fig:travelhtn} shows some simplified methods and their associated sub/tasks; the actual methods and tasks in the system's code differs slightly from this simple form.
For the methods in this system, no preconditions are specified.  In addition to the $(head, task, subtask)$, methods in \systemname{} also include details for the \verify step.

As an example, in Figure~\ref{fig:travelhtn}, the most abstract task \taskname{process user request} decomposes into four subtasks (Lines 3-6) and the \verify context includes additional details (Lines 7-9).  The second subtask \taskname{choose departing flight} is further decomposed by \methodname{m-choose-departing-flight} (Lines 10-17) where one of the subtasks is to understand the flight tool, as described in \methodname{m-understand-flight-tool} (Lines 18-22).

\simplepar{Workflow of \systemname{}}
Figure~\ref{fig:system-workflow} shows an overview of the system with the key components of the system.  Along the top of the diagram, moving left to right, are the agentic components that produce an action. 
The bottom of the diagram shows the components that make up the environment.

Algorithm~\ref{alg:procllm} provides pseudocode for \systemname{}.
If methods \methods{} are provided, Line~\ref{line:updateTask} calls \updateTask to decompose \tasks{}.
\updateTask (Lines~\ref{line:update-tasks-start}-\ref{line:update-tasks-end}) decomposes $head(\tasks{})$ if there are relevant methods.
If \verified is True, \updateTask will remove $head(\tasks)$ and may further decompose \tasks{}, where the leaves are primitive tasks that describe \emph{what} \actionLLM should do.  
\tasks{} remains unchanged if \verified is False, \methods{} is empty, or there are no relevant methods for $head(\tasks{})$. 

After calling \updateTask, \systemname{} then processes a sequence of tasks $\tasks{}$ up to the horizon \Horizon (Line~\ref{line:until-horizon}) or until \tasks{} is empty (Line~\ref{line:empty-task}).
Line~\ref{line:action-context} assembles an action context $\contextAction$ from the current environment state $\mdpstate{}$, the execution $\trace$ of the last system execution, and the current task $\task{}$. 
\contextAction is processed by \actionLLM  (Line~\ref{line:action-llm}) which produces 
a parameterized action \mdpaction{}, which is either an internal \verify action or an external read/write/append action.  
For a \verify action (Lines~\ref{line:verify-start}-\ref{line:verify-end}), Line~\ref{line:verify-context} assembles a verify context \contextVerify  from the system state and current task, which is processed by the \verifyLLM (Line~\ref{line:verify-llm}). 

If there remain tasks to complete, Lines~\ref{line:external-start}-\ref{line:external-end} execute external actions in the environment.  
These external actions are  JSON output that is parsed into an action $\mdpaction$. 
For this paper, 
a set of python scripts implements the transition $\mdptransition(\mdpstate, \mdpaction)$), which produces a new system state $\mdpaction$ that is fed into the agent for the next cycle. 
\frommak{cosmetic: if there is space in the paper, it might be worth going into a little bit of detail about how Lines 16-21 work}

\section{Empirical Evaluation}
\label{sec:evaluation}
We evaluate \systemname{} on two benchmarks with external APIs and two with a combinatorial solution space often solved with search, summarized in the table below.
\textbf{Travel Planning (TP)}, based on \cite{xie2024travelplanner}, requires booking a single-city itinerary using several tools.
\textbf{Recipe Generator (RG)}, a synthetic problem, requires proposing a recipe from a list of ingredients using a tool and solver.
\textbf{Blocks World (BW)}, a planning benchmark \cite{valmeekam2023planbench}, requires rearranging stacks of blocks.
\textbf{Unit Movement (UM)}, another synthetic benchmark, requires moving units around a graph.
    \begin{center}
    \begin{small}
    \begin{tabular}{llllll}
     & \solverpy? & \toolsFile? & Problem Type & Source  \\ 
    \toprule
    TP  & Y & Y  & Tool Use &  \cite{xie2024travelplanner} \\
    RG & Y & Y &  Algorithmic Tool Use & This paper \\
    BW  &  &  & Planning & \cite{valmeekam2023planbench} \\
    UM  &  &  & Game & This paper\\
    \bottomrule
    \end{tabular}
    \end{small}
    \end{center}
~\\
The agent is given a horizon of $\Horizon=100$.
Solutions are automatically verified using human-written simulators and test functions. 
We test three conditions: human-created \methods{} (\humanTN), LLM-created \methods{} (\llmTN), or $\methods = \emptyset$ (\noTN).
For \noTN, \tasks{} has one task describing the end conditions for solving the problem instance; it functions similar to a vanilla reflexion agent \cite{shinn2023reflexion}. 
Experiments are performed on an AMD EPYC 7H12 64 core CPU with the LLMs being served through ollama hosted on A100 GPUs.

\simplepar{Travel Planning}
This tool-use problem is based on the TravelPlanner benchmark \cite{xie2024travelplanner}, where the agent can access database APIs (\texttt{FlightSearch}, \texttt{RestaurantSearch}, etc.) to create an itinerary satisfying a user request. 
The original benchmark contains very difficult multi-constraint tasks where GPT-4 had a low $0.6\%$ success rate \cite{xie2024travelplanner}. 
Although other works have claimed higher success rates on this benchmark, they require considerable human-written code scaffolding \cite{hao2024large} or they use a simplified version of this benchmark (e.g., eliminating tool use in the case of \cite{gundawar2024robust}). 

We use several simplifications so task completion is within reach of current models while keeping success non-trivial. 
First, we filter the validation set of \cite{xie2024travelplanner} to obtain single destination travel requests.
Second, we used a subset of the original benchmark: book a flight from  origin to destination, book accommodation, and book a return flight. 
This subtask benchmark remains sufficiently difficult.
\requestFile is filled with a user request from the validation set of \cite{xie2024travelplanner}.  
\problemSpecFile specifies that the agent should only fulfill parts of the request in order (in contrast to \cite{xie2024travelplanner}). 

We created 43 problems to solve for TP.
Table \ref{tab:travel} shows the success rate for completing the task \emph{up to a given column}. 
For example, for \texttt{Flight 2}, Nemotron with \humanTN chose correct departing and returning flights in 41.9\% of the requests. For both Nemotron and GPT-oss, \humanTN substantially improves task success rate over \noTN. 

\begin{table}
  \caption{Comparison: Mean success rate on subtasks in reduced Travel Planner benchmark (Verification set from \cite{xie2024travelplanner})}
  \small\centering
\begin{tabular}{lllll}
$Model$ & Flight 1 & Flight 2 & Hotel  \\ \toprule
Nemotron 70b \noTN & 0.023 & 0.0  & 0.0 \\
Nemotron 70b \humanTN & 0.535 & 0.419 &  0.116 \\
GPT-oss 120b \noTN & 0.0 & 0.0 & 0.0  \\
GPT-oss 120b \humanTN & 0.814 & 0.605 & \textbf{0.395}
\\\bottomrule
\end{tabular}
\vspace{-1em}
\label{tab:travel}
\end{table}

\simplepar{Recipe Generator (RG)}
In this tool use task, an agent is given a list of cooking ingredients and asked to return a dish that can be made from these ingredients. To process user requests, the agent uses two python functions:
    \taskname{get\_ingredient\_from\_dish(dish)} returns the ingredients for a dish, and 
    \taskname{get\_dish\_from\_ingredient(ingredient)} returns a list of dishes for an ingredient.
\verify succeeds for any dish that can be made from any subset of ingredients, and the agent can write arbitrary python code calling these functions.

Using GPT-4, we generated a database of ingredients and dishes.
A problem instance is constructed by randomly selecting a dish, extracting its ingredients, and augmenting this list with additional ingredients. 
This process ensures solvable problems that are also challenging.

Over 20 problem instances (not shown for space reasons), a task network improves results.
GPT-oss-120b had a base success rate of 0\% for \noTN and 25\% for \humanTN.
Because solutions to this task may require algorithmic solutions (e.g. filtering api call results), we applied a coding fine-tuned LLM Nemotron-70b, which averaged a success rate of 50\% for \noTN and 80\% for \humanTN. 

\simplepar{Blocks World (BW)}
This problem adapts the BlocksWorld benchmark from PlanBench \cite{valmeekam2023planbench}, where GPT-4 succeeded on $34.3\%$ problems. 
In PlanBench, problems have 3-5 blocks, and the LLM  prompt includes the rules of the task, an example problem, and a solution to the example problem.

We made several changes. 
\problemSpecFile describes the problem and the acceptable solution format and \requestFile contains the initial state. 
However, these files provide \emph{no explicit solution example}.
To assess the impact of problem complexity, we randomly choose $b$ initial blocks that must be in a final stack of height $h$; a solution is correct for stacking $h$, regardless of other blocks $b-h$.

\begin{figure}
     \begin{center}
     \includegraphics[width=0.450\textwidth]{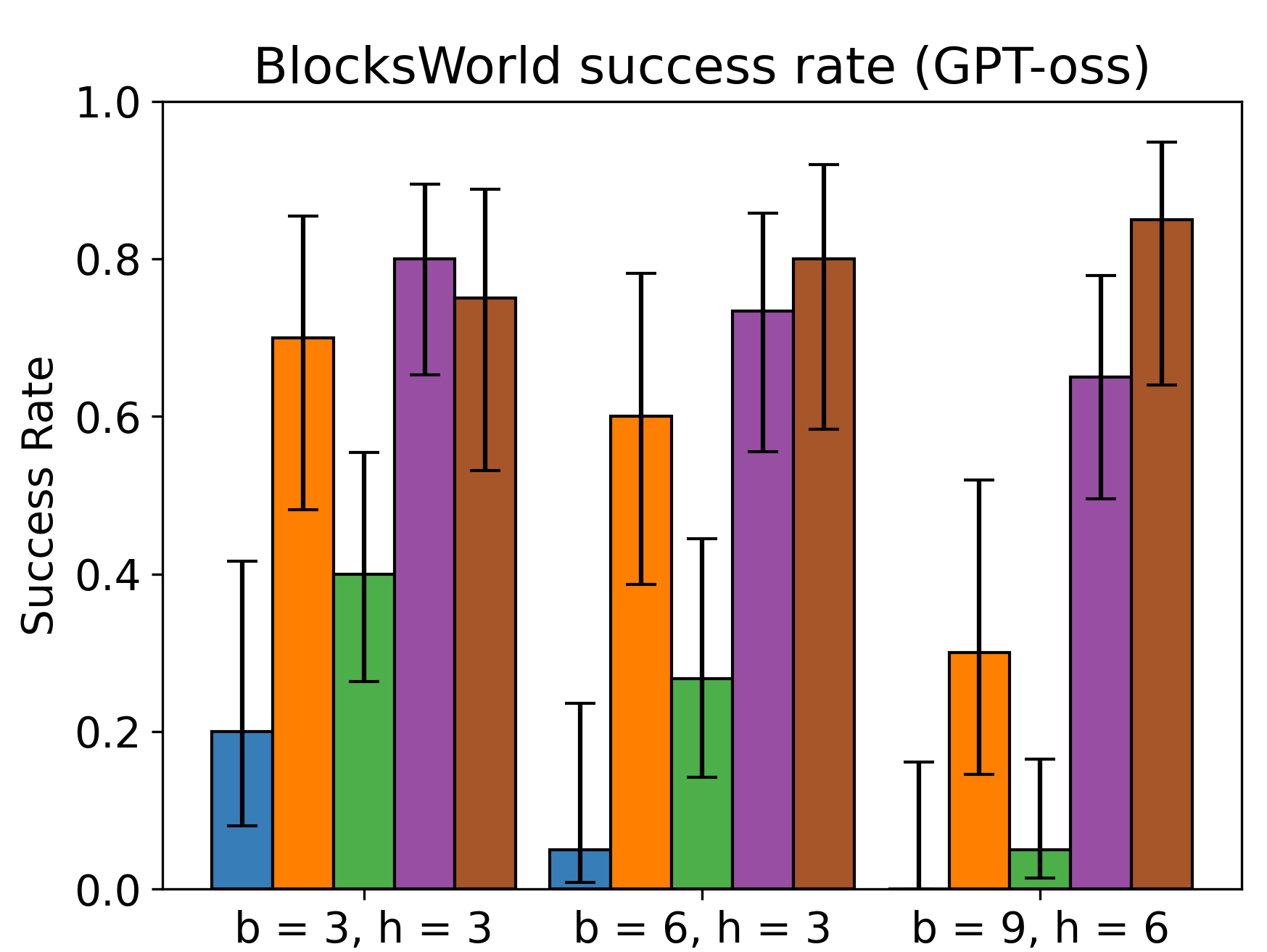}
     \\
     \vspace{0.4em}
     \includegraphics[width=0.450\textwidth]{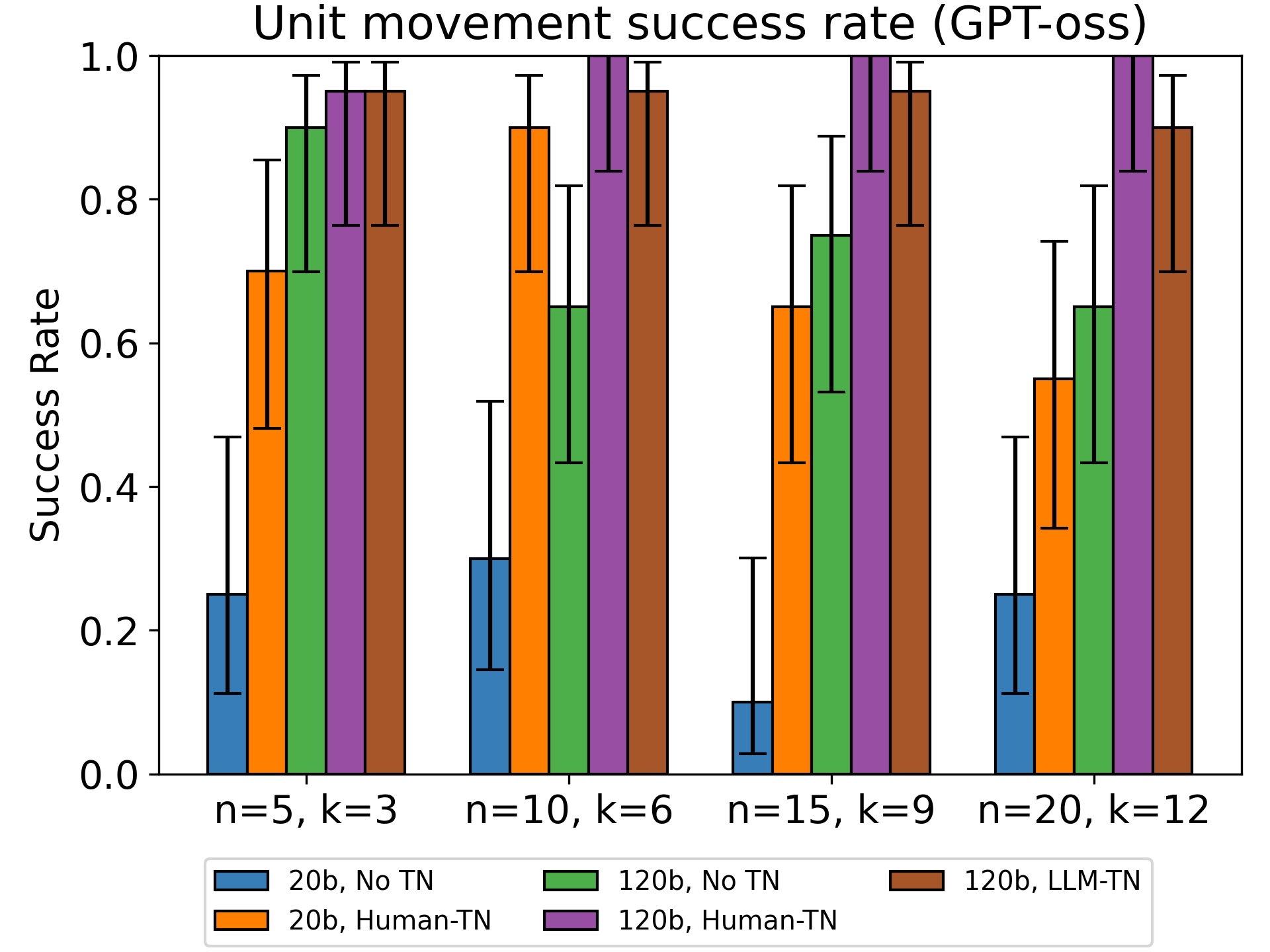}
     \end{center}
     \caption{Success rates for BlockWorld and Unit Movement domains, evaluated across increasing problem complexity ($b$ starting blocks and $h$ final stack height for BW, number of units $n$ for UM). Bars show mean success rates, with 95\% Wilson confidence intervals. 
     }
     \label{fig:blockresults}
 \end{figure}

Fig \ref{fig:blockresults} reveals a downward trend in task success as complexity increases (from $b=3$ to $b=9$). 
Either task network (\humanTN or \llmTN) greatly increases the success rate; the success rate of GPT-oss-120b \humanTN remains 70\% in contrast to GPT-oss-120b \noTN of below 5\%. 
Furthermore, the smaller GPT-oss-20b \humanTN significantly outperforms GPT-oss-120b.

\simplepar{Unit Movement (UM)}
In this task, for each problem instance we generate a graph $G$ with the following properties (Figure~\ref{fig:unitmovementproblem} shows an example problem).
A \textbf{target node} $v_T \in G$ is adjacent to \textbf{four neighbors} $v^{adj}$, each with an edge $(v^{adj}, v_T)$.
Each neighbor has three \textbf{outer neighbors}; that is, for each node $v^{adj}_i$, there are at least 3 outer nodes $v^{out}$ with an edge $(v^{adj}_i, v^{out})$.
 After generating the above node sets, we add 12 more random edges between neighboring nodes (excluding the target $v_T$).
For three of the sections of the graph we place $n$ units to random nodes in each section.  
We randomly name each node (e.g. Seaside, Canyon, Gardens) and each unit based on their section (e.g. Bravo\_0, Alpha\_1). 
The goal of the task is to move units so that the target is surrounded from three neighbors with at least least $k$ units at each node.

     

We varied problem complexity by the number of $n$ units per group and the number $k$ units required for surrounding. 
Figure~\ref{fig:blockresults} reveals two insights. First, \humanTN increases the task success rate regardless of the base capability of the LLM used in the system; at  ($n=10,k=6$), the 20b model \humanTN outperforms 120b \noTN. Second, higher $(n,k)$ values seem show a non-linear correlation with task difficulty; future work should explore whether the LLM is  exploiting problem structure to solve larger problems.

\section{Discussion}
The evaluation of \systemname{} shows that \humanTN and \llmTN improve task success rate, often significantly and even across \noTN success capability and different benchmarks. 
We also varied problem complexity and showed that, in BW and UM, the LLM system with a task planner seems to have a slower decrease in task success rate than the LLM system without a task planner. 
This is true in both domains for GPT-oss-120b, but the results for GPT-oss-20b show less improvement; we hypothesize that smaller models such as GPT-oss-20b can run into context length issues that are not addressed by our task network approach.
We further examine several more questions about these results.

\simpleparNoEnd{Why is \noTN doing so poorly?}
Although not obvious from the results presented so far, there are some subtle differences in model behavior that we would like to discuss. To explain, consider the first few steps that an LLM may need to take to solve one of the agentic problems seen so far. In each of the problems, there is a \texttt{request.txt} and a \texttt{problem specification.txt}. The first few steps of solving an agentic task is to read each of these files and then take notes on the files into the short-term memory file (\notesFile). Some models (e.g., nemotron, codestral, llama) are naturally proficient in this and will generate a sequence of actions:
\begin{small}
    \begin{enumerate}[noitemsep,topsep=0pt,parsep=0pt,partopsep=0pt]
        \item Read \texttt{request.txt}
        \item Append [request details]
        \item Read \texttt{problem specification.txt}
        \item Append [spec details]
    \end{enumerate}
\end{small}
This sequence of actions can be generated even if the task network does not have explicit task nodes about taking notes (or even without a task network at all). However, our experiments with GPT-oss-120b revealed the system would get stuck in loops like:
\begin{small}
    \begin{enumerate}[noitemsep,topsep=0pt,parsep=0pt,partopsep=0pt]
        \item Read \texttt{request.txt}
        \item Read \texttt{problem specification.txt}
        \item Read \texttt{request.txt}
        \item Read \texttt{problem specification.txt}
        \item etc.
    \end{enumerate}
\end{small}
This causes the system to timeout and fail as a result. We can alleviate this problem for domains such as BW by explicitly extending the final goal for the task. For example, we could change \taskname{solve the user request}  into \taskname{solve the user request and have notes on the request and problem specification}. But such\\ changes may not scale if a problem has many goals.

 \simpleparNoEnd{Are LLM-constructed task networks better? (\llmTN)}
To study \llmTN, we constructed a prompt that includes the problem specification, the agent prompt, and two valid task network node examples and asked Gemini 2.5 Pro to produce a task network (LLM-TN). In Figure~\ref{fig:blockresults}, we see that task success rate of \llmTN (brown bars) does better than \noTN while varying against \humanTN.

\begin{figure}
    \begin{center}
     \includegraphics[width=0.6\textwidth]
     {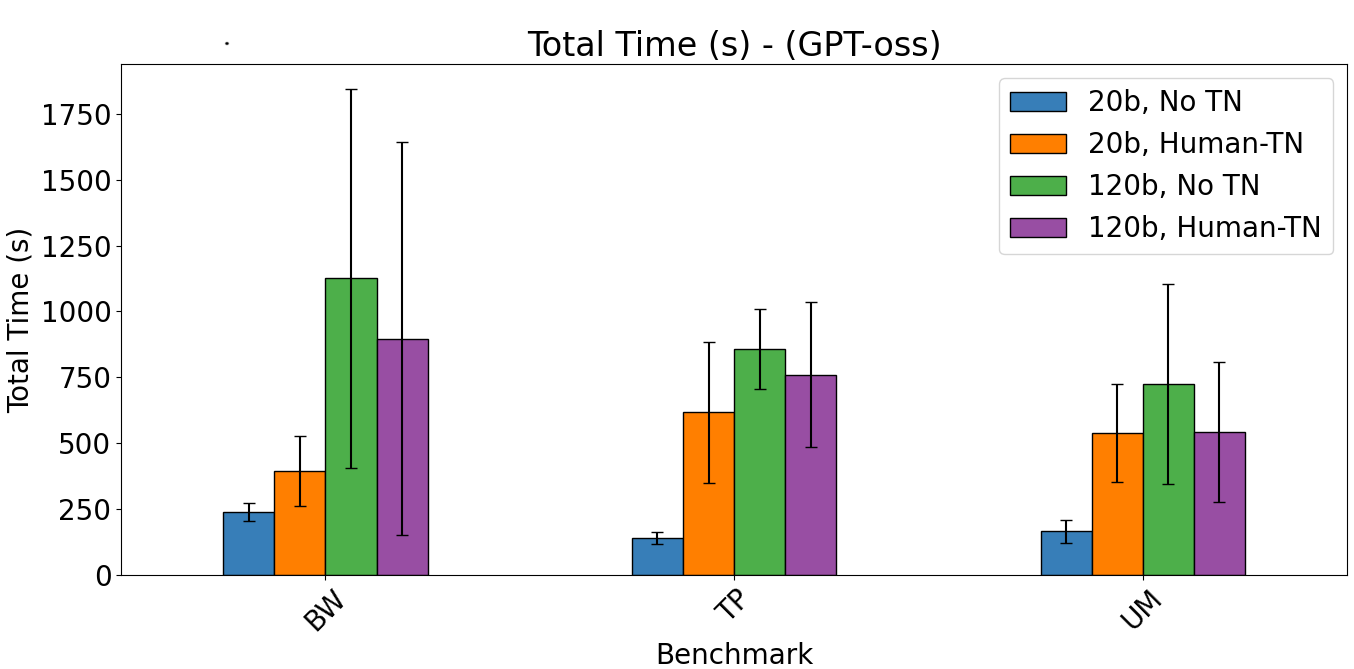}
     \end{center}
     \caption{A comparison of average runtime statistics for GPT-oss across different benchmarks (BW - BlocksWorld, TP - Travel Planner, UM - Unit Movement). 
     (See Fig.~\ref{fig:timeresults-detailed} for full details) 
     \frommak{Add iteration plots if there is space. }
     }
     \label{fig:timeresults}
 \end{figure}

\simpleparNoEnd{Does \humanTN or \llmTN save runtime?}
Figure~\ref{fig:timeresults} shows runtime statistics for GPT-oss-120b in \systemname{} across the evaluation benchmarks (Figure~\ref{fig:timeresults-detailed} shows  full details). For Blocks World and Unit Movement, the statistics are averaged over all problem sizes.

In three of the four benchmarks, the task planner decreases the number of system iterations required for a given problem instance. The exception is Recipe Generator, where the LLM with \noTN produces quick but always incorrect outputs, unlike other benchmarks where \noTN typically fails via timeout. One explanation is that the LLM can treat the request in Recipe Generator as a query to be answered directly, bypassing the APIs altogether. This shortcutting behavior mirrors earlier findings (\cite{belcak2025universal}) that LLMs can default to superficial strategies unless constrained externally. 

Using a task network typically increases the average time per iteration while decreasing the total number of iterations. One important takeaway is that using the task network does not significantly increase the time it takes to complete a task. In some cases, such as on BlocksWorld, using GPT-oss-20b in \systemname{} will outcompete GPT-oss-120b in both time and performance metrics. 

\section{Conclusion}
We have argued that procedural knowledge helps LLM workflows. 
We formalized, implemented, and evaluated \systemname{} that uses HTNs to improve LLM performance on complex agentic tasks. 
\systemname{} leveraged an HTN to decompose complex tasks into subtasks that completed by an agentic LLM.  
While \systemname{} uses a file system as the environment, this framework could be applied to employ LLM agents in more general MDP domains (e.g., games, control problems). 

We demonstrated on four benchmarks that the HTN decomposition in \systemname{} significantly increased task success in difficult multi-step agentic tasks. This is effective regardless of the base LLM. 
This improvement uses natural language descriptions; \systemname{} does  not require translation to a planning formalism (e.g., PDDL, HDDL).
Using natural language allows a flexible and natural tradeoff in the level of abstraction for HTN planning.
The high-level planning afforded by the task planner increases the likelihood of completing long, complex tasks and also reduces the probability of many common agentic LLM issues, such as action looping, that are exacerbated by a deficiency in implicit task planning.

LLM generated task networks sometimes do well, but they seem to generally lack the specificity of human generated networks. 
In future work, it would be interesting to see if LLMs can build task networks from the ground up in a curriculum learning fashion, starting with simple tasks that gradually become more complex.  Also, it would be interesting to explore an equivalent approach in \cite{belcak2025universal} where an LLM transforms the HTN recipe into code for speed and consistent replicability.
Finally, procedures are frequently codified into documents as recipes, checklists, or standard operating procedures. Beyond learning networks through curriculum learning, we could explore methods for using RAG (e.g., \cite{shi2024compressing}) or fine-tuning (e.g., \cite{vermaEtal25.arxiv.teachingLLMs}) to improve the distillation of task networks from domain documents

\section*{Acknowledgments}
We thank NRL for funding this research.


\clearpage
\bibliographystyle{ACM-Reference-Format} 
\bibliography{aaai2026}


\newpage
\onecolumn

\appendix

\renewcommand\thefigure{\thesection.\arabic{figure}} 
\setcounter{figure}{0}

\section{Additional Figures}
\label{app:problem-details}

\begin{figure}
     \begin{center}
     \includegraphics[width=0.50\textwidth]{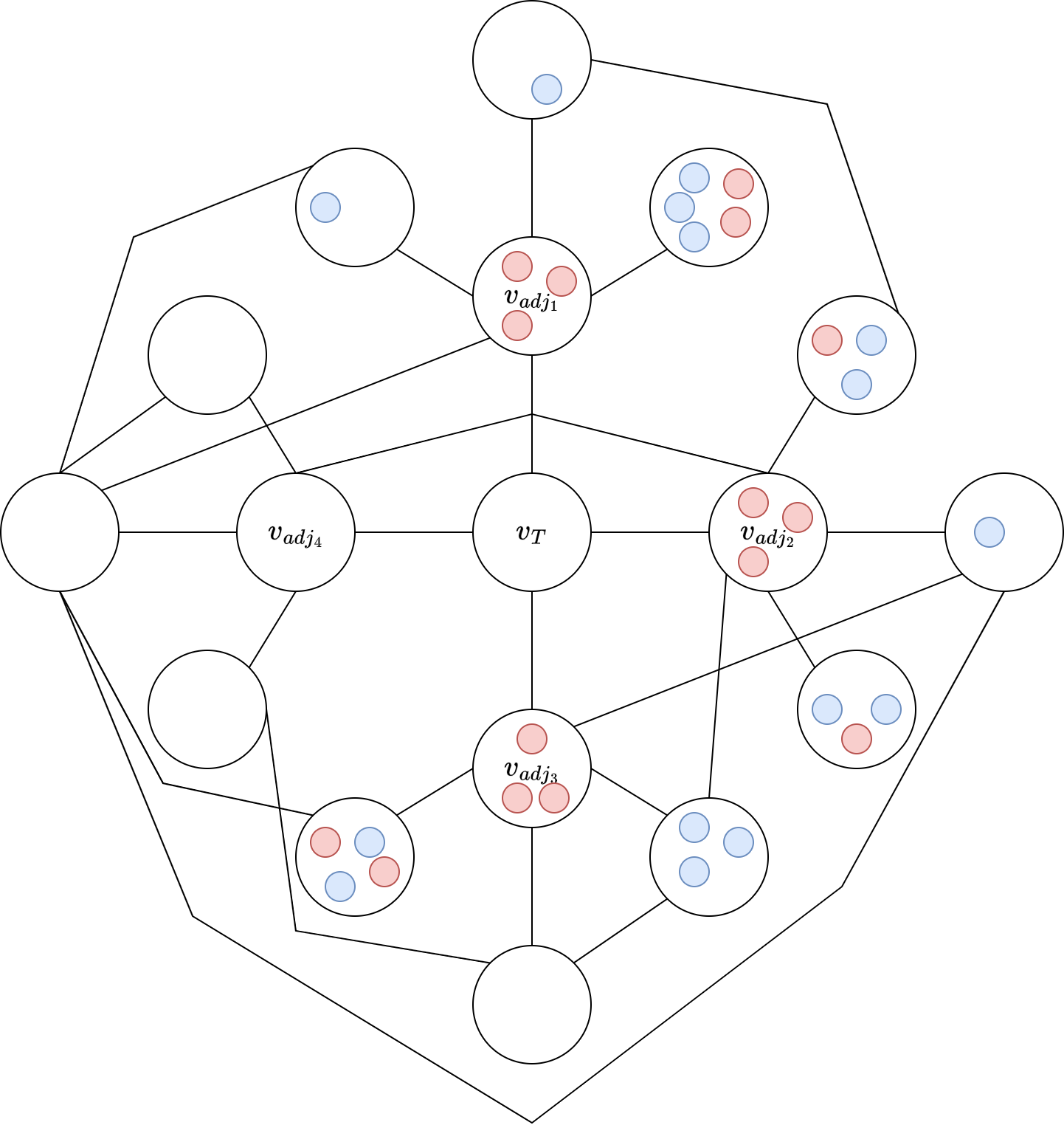}
     \end{center}
     \caption{An example problem instance for the unit movement domain showing the initial starting position (blue circles) and a valid solution (red circles).  
     \frommak{Is it possible to color the edges of each layer?  That might help with the explanation.}
     }
     \label{fig:unitmovementproblem}
 \end{figure}

\newpage
\label{app:eval}

\begin{figure}
     \begin{center}
     \includegraphics[width=0.7\textwidth]{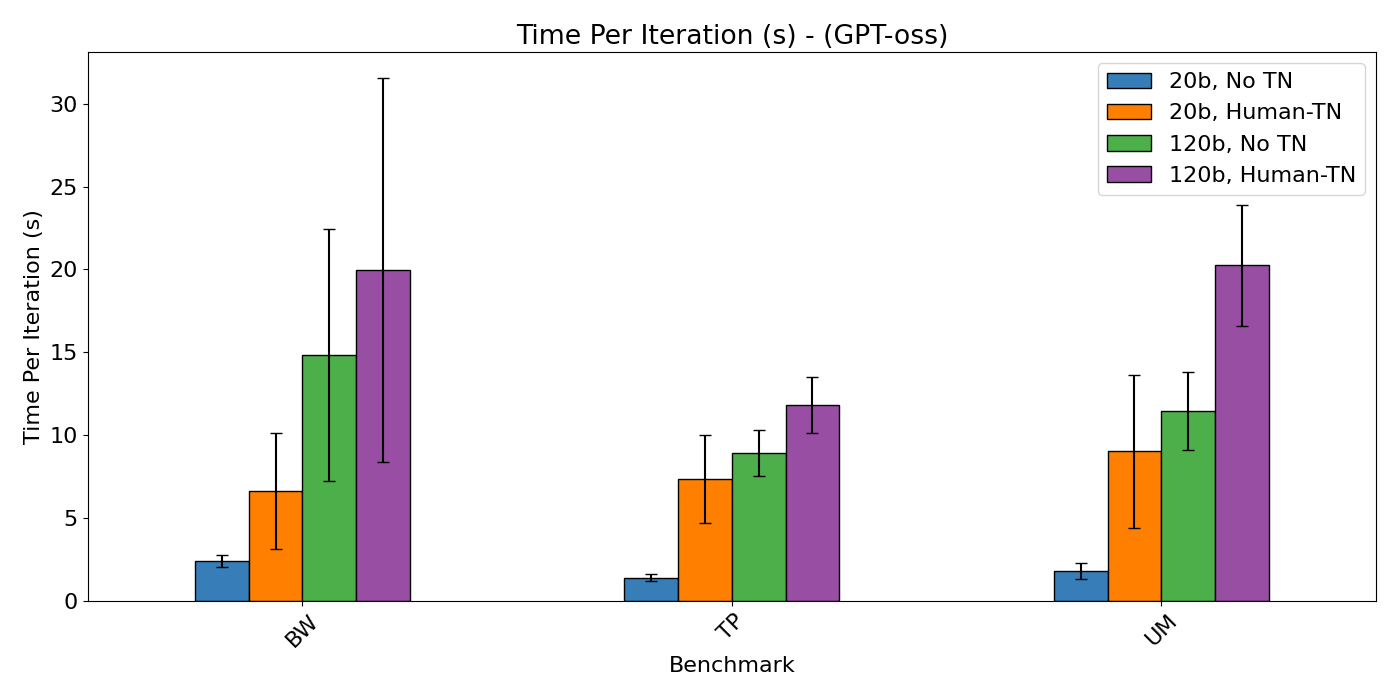}
     \includegraphics[width=0.7\textwidth]{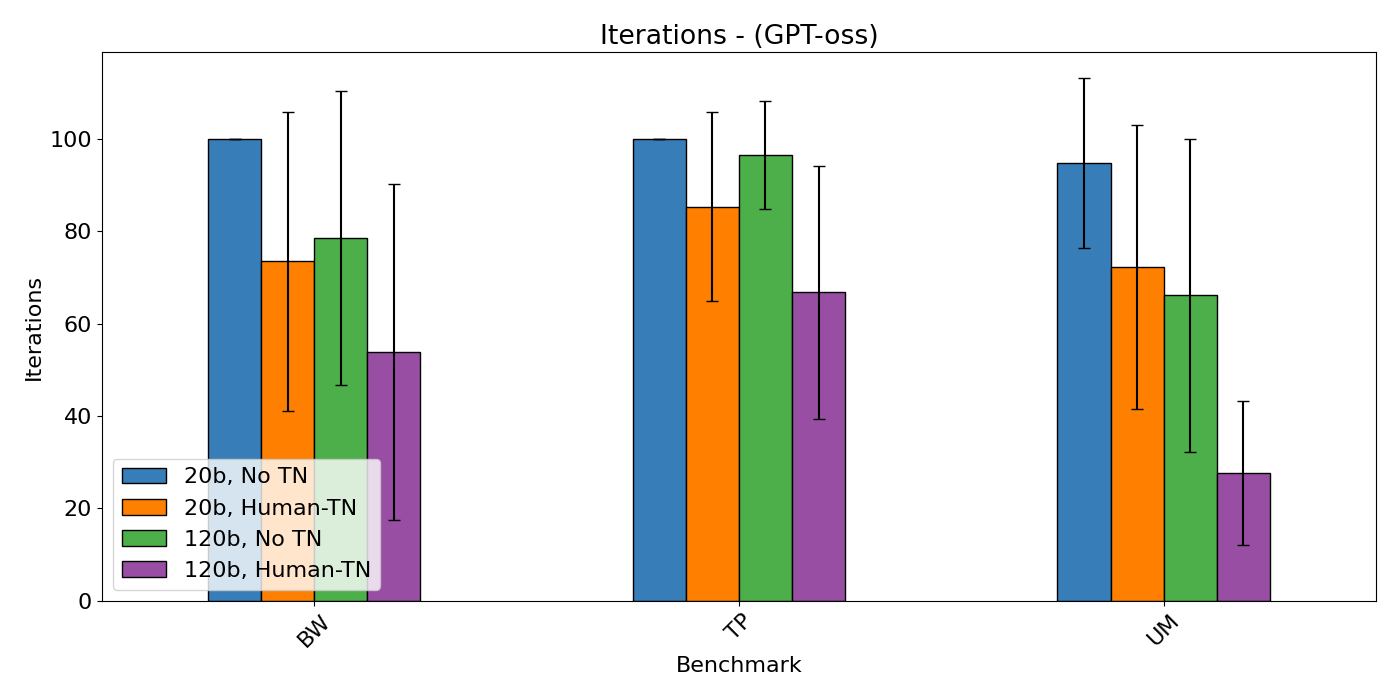}
     \includegraphics[width=0.7\textwidth]{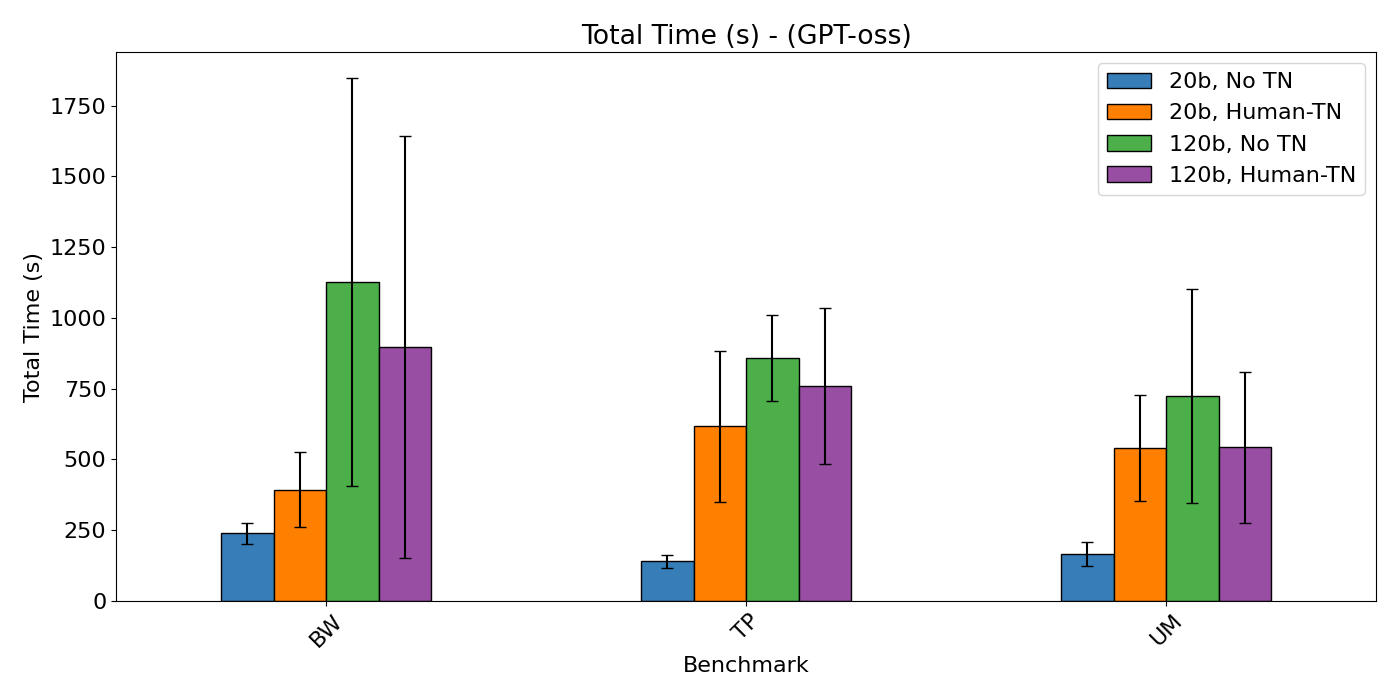}
     \end{center}
     
     \caption{A comparison of average runtime statistics for GPT-oss 120b across different benchmarks (BW - BlocksWorld, TP - Travel Planner, UM - Unit Movement). \frommak{I really like the detail of this plot but worry we don't have room.  Maybe this larger plot could move to the appendix and be replaced with a stacked bar plot that is 30mm high? }}
     \label{fig:timeresults-detailed}
 \end{figure}

\newpage
\section{Prompts}
\label{app:prompts}

\subsection{Standard Agent Prompt}
\frommak{I have started using the listings package for these -- it is much easier to make consistent and make changes.  I'll add for these once we get to the cosmetic phase of the paper.}

This prompt is given to the agent at each iteration for all domains. Only the content of the read and writeable files may differ across domains. 
\begin{cvbox}
\begin{Verbatim}[breaklines=true, samepage=false]
Your goal is to interact with a system that processes commands in order to solve a given task. The task is specified in the problem specification document and the task instance in the request file. You MUST read and take notes on these files to solve the problem. You will only be able to see the information from the immediately preceding action, so make sure you take notes such that this information is not lost. 
You have the following actions and files available to you:

1. Read: reads [action_arg1]. Line numbers will be displayed.
Read [action_arg1]

2. Write: overwrites text file [action_arg1] with [action_arg2].
Write [action_arg1] [action_arg2]

3. Append: appends the text file [action_arg1] with [action_arg2].
Append [action_arg1] [action_arg2]

4. Verify: Send a request to check if current task has been achieved.
Verify

You are allowed to read and write the following file (r/w):
solver.py (r/w): this file is automatically executed after every revision. The standard output of this file is written to output.txt
files/notes.txt (r/w): the contents of this file will always be displayed in the notes section down below. 
answer.txt (r/w): a file to hold the final answer to user request

You are allowed to access the following read-only files (r):
files/request.txt (r): the user request
files/problem_specification.txt (r): a problem specification document
files/tools_specification.txt (r): a document specifying tools
output.txt (r): the stdout of solver.py

The contents of files/notes.txt is:
***NOTES START***
{}
***NOTES END***

Your last response was:
***LAST RESPONSE START***
{}
***LAST RESPONSE END***

The last ten commands executed were:
***LAST COMMAND START***
{}
***LAST COMMAND END***

Here is the output of your last command:
***LAST COMMAND OUTPUT***
{}
***LAST COMMAND OUTPUT END***

Your Limitations:
1. Do not attempt to read or write to any files not listed. This will result in a file access denied error.
2. You can only open one file at a time.
3. If you get a file access denied error, make sure you use the correct path and extension.

General Instructions:
1. Your response should be in a json format (formatted with newlines) and include the following components: 
 * observation: a detailed observation about the current state of the task and what you have done so far
 * thought: reasoning about what next steps are valid and how to proceed
 * action {{name: , action_arg1: , action_arg2: }}: action [read, write, append, or verify] and arguments for the action
2. There is no limit on how long your json can be, but do not include additional nested components.
3. If you receive a JSONDecode error, this means your json response is improperly formatted.
4. Think step by step. Justify each step with something from existing notes or previous responses. 
5. Respond only with your json action. Use an empty string if any arguments are unneeded. 
6. A task planner will give you a task to follow with an effect. Focus on achieving this effect. 
7. Call the verify action when the current effect is satisfied to move onto the next task.
8. You MUST take notes to solve the problem. Make sure to take notes after a read.
9. Do not include section headings (e.g. ***) in your response.

Your current task is:
Task: {}
The effect that you are trying to achieve is:
Effect: {}
\end{Verbatim}
\end{cvbox}

\subsection{Agent Action Examples}
\label{app:action-examples}

The agent is expected to respond with a JSON that will be parsed and processed as an action. Here are some examples:
\begin{cvbox}
\begin{Verbatim}[breaklines=true, samepage=false]
{
  "observation": "Currently, no notes have been taken and no files have been read. The task is to take notes on the problem specification file.",
  "thought": "To achieve the effect of having notes contain the information from the problem specification file, I first need to read the problem specification file. Then, I will append the contents of this file to the notes.txt file to ensure persistence of the information.", 
  "action": {
    "name": "Read",
    "action_arg1": "files/problem_specification.txt",
    "action_arg2": ""
  }
}
\end{Verbatim}
\end{cvbox}

\begin{cvbox}
\begin{Verbatim}[breaklines=true, samepage=false]
{
  "observation": "Having read the request file (files/request.txt), I've obtained additional context: a 3-day trip from Missoula to Dallas (March 23rd-25th, 2022) with a budget of $1,900. This aligns with my initial notes but adds a budget constraint. My current understanding is that I need to use Flights.py for flights and Accommodations.py for lodging within this budget. However, I still lack detailed insights from the problem specification file beyond the initial notes. To ensure completeness, I'll append these new details to notes.txt.",

  "thought": "With the request's context in mind, particularly the budget, my next steps should involve utilizing the specified tools (Flights.py and Accommodations.py) while considering the $1,900 budget. Before executing code, it's prudent to update my notes with the latest understanding, including the budget constraint, to maintain a comprehensive record of my approach.",

  "action": {
    "name": "Append",
    "action_arg1": "files/notes.txt",
    "action_arg2": "Additional context from request file: Budget for Missoula to Dallas trip (March 23rd-25th, 2022) is $1,900. Next steps involve using Flights.py and Accommodations.py with budget consideration."
  }
}

\end{Verbatim}
\end{cvbox}

\begin{cvbox}
\begin{Verbatim}[breaklines=true, samepage=false]
{
  "observation": "Read the problem specification which defines the allowed actions and required output format. Notes provide the initial state, goal state, and a sequence to unstack all blocks to the table. To achieve the goal, we will first execute the unstack/put sequence to clear the table, then pick and stack blocks in the order required by the goal: cyan at bottom, then blue, orange, green, yellow, purple. This yields a complete plan using only pick, put, unstack, and stack actions.",
  "thought": "Construct a plan consisting of the 8 unstack/put steps from the notes, followed by pick/stack steps to build the goal tower. Write this plan to answer.txt in the exact line-by-line format required.",
  "action": {
    "name": "Write",
    "action_arg1": "answer.txt",
    "action_arg2": "unstack black green\nput black\nunstack green cyan\nput green\nunstack cyan blue\nput cyan\nunstack blue purple\nput blue\npick blue\nstack blue cyan\npick orange\nstack orange blue\npick green\nstack green orange\npick yellow\nstack yellow green\npick purple\nstack purple yellow"
  }
}
\end{Verbatim}
\end{cvbox}

\subsection{Example mid-iteration prompt}

\begin{cvbox}
\begin{Verbatim}[breaklines=true, samepage=false]
Your goal is to interact with a system that processes commands in order to solve a given task. The task is specified in the problem specification document and the task instance in the request file. You MUST read and take notes on these files to solve the problem. You will only be able to see the information from the immediately preceding action, so make sure you take notes such that this information is not lost. You have the following actions and files available to you:

1. Read: reads [action_arg1]. Line numbers will be displayed.
Read [action_arg1]

2. Write: overwrites text file [action_arg1] with [action_arg2].
Write [action_arg1] [action_arg2]

3. Append: appends the text file [action_arg1] with [action_arg2].
Append [action_arg1] [action_arg2]

4. Verify: Send a request to check if current task has been achieved.
Verify

You are allowed to read and write the following file (r/w):
solver.py (r/w): this file is automatically executed after every revision. The standard output of this file is written to output.txt
files/notes.txt (r/w): the contents of this file will always be displayed in the notes section down below. 
answer.txt (r/w): a file to hold the final answer to user request

You are allowed to access the following read-only files (r):
files/request.txt (r): the user request
files/problem_specification.txt (r): a problem specification document
files/tools_specification.txt (r): a document specifying tools
output.txt (r): the stdout of solver.py

The contents of files/notes.txt is:
***NOTES START***
The user will provide a request to make a travel plan. You do not need to complete the entire travel plan, but only the flight and accommodations for the user's trip. You will need to use code files Flights.py and Accommodations.py to access the database for required to solve the user request

Your solution should be returned in a json format in answer.txt with two flights (departing and returning) as well as the accommodation used for the duration of the traveler's stay:

departing_flight: flight number
returning_flight: flight number
accommodation: name

Be aware that the accommodations returned by the Accommodations tool may have names that do not seem to correspond with the city that you used for the search; this is expected behavior and you should proceed with the results you obtained using the tools.

Origin: Detroit
Destination: San Diego
Departure Date: 2022-03-05
Returning Date: 2022-03-07
Duration: 3 days
Budget: $3000
Traveler: solo


FlightSearch.run method:
- Inputs:
  * Departure City (string)
  * Destination City (string)
  * Date (string in YYYY-MM-DD format)
- Expected Output:
  * List of flight options (each containing flight number, departure/arrival times, price, etc.)

***NOTES END***

Your last response was:
***LAST RESPONSE START***
{
  "observation": "We have read the tools specification which shows how to import and use the Flights tool. To obtain flight options we need to write code in solver.py that imports the Flights class, creates an instance, runs the search for Detroit to San Diego on 2022-03-05, and prints the results. The solver will then be executed and its stdout captured in output.txt for later analysis.",
  "thought": "Next step is to write the described code into solver.py so that the flight search is performed and results are printed. This will allow us to later read output.txt and extract a departing flight number and cost.",
  "action": {
    "name": "Write",
    "action_arg1": "solver.py",
    "action_arg2": "from tools.Flights import Flights\n\ndef main():\n    FlightSearch = Flights()\n    results = FlightSearch.run(\"Detroit\", \"San Diego\", \"2022-03-05\")\n    print(results)\n\nif __name__ == \"__main__\":\n    main()\n"
  }
}
***LAST RESPONSE END***

The last ten commands executed were:
***LAST COMMAND START***
Verify 
Read files/request.txt
Append files/notes.txt
Verify 
Read files/tools_specification.txt
Append files/notes.txt
Verify 
Read solver.py
Read files/tools_specification.txt
Write solver.py
***LAST COMMAND END***

Here is the output of your last command:
***LAST COMMAND OUTPUT***
Updated solver.py:
from tools.Flights import Flights

def main():
    FlightSearch = Flights()
    results = FlightSearch.run("Detroit", "San Diego", "2022-03-05")
    print(results)

if __name__ == "__main__":
    main()

Code executed with stdout:
        Flight Number  Price DepTime  ... OriginCityName DestCityName Distance
3306155      F3528556    720   08:33  ...        Detroit    San Diego   1956.0

[1 rows x 9 columns]

***LAST COMMAND OUTPUT END***

Your Limitations:
1. Do not attempt to read or write to any files not listed. This will result in a file access denied error.
2. You can only open one file at a time.
3. If you get a file access denied error, make sure you use the correct path and extension.

General Instructions:
1. Your response should be in a json format (formatted with newlines) and include the following components: 
 * observation: a detailed observation about the current state of the task and what you have done so far
 * thought: reasoning about what next steps are valid and how to proceed
 * action {name: , action_arg1: , action_arg2: }: action [read, write, append, or verify] and arguments for the action
2. There is no limit on how long your json can be, but do not include additional nested components.
3. If you receive a JSONDecode error, this means your json response is improperly formatted.
4. Think step by step. Justify each step with something from existing notes or previous responses. 
5. Respond only with your json action. Use an empty string if any arguments are unneeded. 
6. A task planner will give you a task to follow with an effect. Focus on achieving this effect. 
7. Call the verify action when the current effect is satisfied to move onto the next task.
8. You MUST take notes to solve the problem. Make sure to take notes after a read.
9. Do not include section headings (e.g. ***) in your response.

Your current task is:
Task: choose a departing flight
The effect that you are trying to achieve is:
Effect: notes contains details about a flight number and cost for a flight from origin to destination on departure date consistent with request preferences
\end{Verbatim}
\end{cvbox}
\subsubsection{Example response to the previous prompt}
\begin{cvbox}
\begin{Verbatim}[breaklines=true, samepage=false]
{
  "observation": "The solver.py execution printed flight search results to output.txt, showing a flight with number F3528556, price 720, departure time 08:33, etc. We need to capture these details in notes.txt.",
  "thought": "To satisfy the effect, I will read output.txt to obtain the flight information, then append a concise note with the departing flight number and its cost to notes.txt.",
  "action": {
    "name": "Read",
    "action_arg1": "output.txt",
    "action_arg2": ""
  }
}
\end{Verbatim}
\end{cvbox}

\section{Domain Specification}
\label{app:domain-spec}
For each domain, we write problem specification documents that the agent needs to read in order to complete a task (e.g. what they need to actually do to complete the agentic task). Each problem instance also has an automatically generated \texttt{request.txt} that describes the problem details for a specific problem instance.
\subsection{Domain: Travel Planner}
\subsubsection{Problem Specification}
\begin{cvbox}
\begin{Verbatim}[breaklines=true, samepage=false]
The user will provide a request to make a travel plan. You do not need to complete the entire travel plan, but only the flight and accommodations for the user's trip. You will need to use code files Flights.py and Accommodations.py to access the database for required to solve the user request

Your solution should be returned in a json format in answer.txt with two flights (departing and returning) as well as the accommodation used for the duration of the traveler's stay:

departing_flight: flight number
returning_flight: flight number
accommodation: name

Be aware that the accommodations returned by the Accommodations tool may have names that do not seem to correspond with the city that you used for the search; this is expected behavior and you should proceed with the results you obtained using the tools.
\end{Verbatim}
\end{cvbox}
\subsubsection{Example User Request}
\begin{cvbox}
\begin{Verbatim}[breaklines=true, samepage=false]
Please create a travel plan for a 3-day trip from Missoula to Dallas scheduled from March 23rd to March 25th, 2022. The budget for this trip is set at $1,900.
\end{Verbatim}
\end{cvbox}

\subsection{Domain: Recipe Generator}
\subsubsection{Problem Specification}
\begin{cvbox}
\begin{Verbatim}[breaklines=true, samepage=false]
The user will provide a request containing a list of ingredients, your job is to select a valid recipe using those ingredients. 

You do not need to use all ingredients provided by the user. Any recipe that can be made from the ingredients provided by the user is a valid recipe. For example, if the user provides the ingredient list:

Chicken Breast
Tomatoes
Soy Sauce
Sesame Oil
Brown Sugar
Garlic

The following recipe:

Dish: Chicken and Tomato Stir Fry
Ingredients:
Chicken Breast
Tomatoes
Soy Sauce
Sesame Oil

is valid solution to the user request, and a valid final output is:

Chicken and Tomato Stir Fry
\end{Verbatim}
\end{cvbox}
\subsubsection{Example User Request}
\begin{cvbox}
\begin{Verbatim}[breaklines=true, samepage=false]
Hello, I'd like to request a recipe for the following ingredients: 
Chicken Breast
Tomatoes
Soy Sauce
Sesame Oil
Garlic
Ginger
\end{Verbatim}
\end{cvbox}

\subsection{Domain: Blocks World}
\subsubsection{Problem Specification}
\begin{cvbox}
\begin{Verbatim}[breaklines=true, samepage=false]
I am playing with a set of blocks where I need to arrange the blocks into stacks. Here are the actions I can do:
Pick up a block
Unstack a block from on top of another block
Put down a block
Stack a block on top of another block
I have the following restrictions on my actions:
I can only pick up or unstack one block at a time.
I can only pick up or unstack a block if my hand is empty.
I can only pick up a block if the block is on the table and the block is clear. A block is clear if the block has no other blocks on top of it and if the block is not picked up.
I can only unstack a block from on top of another block if the block I am unstacking was really on top of the other block.
I can only unstack a block from on top of another block if the block I am unstacking is clear.
Once I pick up or unstack a block, I am holding the block.
I can only put down a block that I am holding.
I can only stack a block on top of another block if I am holding the block being stacked.
I can only stack a block on top of another block if the block onto which I am stacking the block is clear.
Once I put down or stack a block, my hand becomes empty.
Once you stack a block on top of a second block, the second block is no longer clear.

Please provide a plan as a list of actions with space deliminated arguments line by line in the following format in answer.txt:
pick [block]: pick a block up
put [block]: put a block down
stack [block1] [block2]: stack block1 on block2
unstack [block1] [block2]: unstack block1 from on top of block2

An example plan:
unstack blue red
put blue
pick red
stack red green

Write an answer to answer.txt. The plan will be checked automatically so please follow the above format. The only actions allowed are [pick, put, stack, unstack].
\end{Verbatim}
\end{cvbox}
\subsubsection{Example User Request $(n=9, k=6)$}
\begin{cvbox}
\begin{Verbatim}[breaklines=true, samepage=false]
As initial conditions I have that:
the blue block is on the table
the gray block is on top of the blue block
the red block is on top of the gray block
the orange block is on top of the red block
the yellow block is on top of the orange block
the black block is on top of the yellow block
the cyan block is clear
the cyan block is on top of the black block
the purple block is clear
the purple block is on the table
the green block is clear
the green block is on the table
My goal is to have that: 
the orange block is on top of the gray block
the blue block is on top of the orange block
the black block is on top of the blue block
the yellow block is on top of the black block
the red block is on top of the yellow block
\end{Verbatim}
\end{cvbox}

\subsection{Domain: Unit Movement}
\subsubsection{Problem Specification}
\begin{cvbox}
\begin{Verbatim}[breaklines=true, samepage=false]
You are an AI strategist tasked with controlling a team of infantry units in a wargame. You are given the current game state and must output a list of actions for your units to execute in a single turn.

Game State:
The game is played on a network. Units have location, and a unit ID.

Example unit:
{"unit_id": "Eagle_0", "location": "Westwood"},

Available Actions:

* Move:
  {"unit_id": "unit_id", "action_type": "move", "location": target_location}
  The unit will move to target_location. 

Output Format:
Each action must be a valid JSON object as described above in a list. Do not include any other text besides the JSON list.

Example Output for two units:
[
  {"unit_id": "Blue_0", "action_type": "move", "location": "Village"},
  {"unit_id": "Blue_1", "action_type": "move", "location": "Southshore"}
]

To make the task easier for more complex problem instances, it can be beneficial to group units together, decide on a general top level movement strategy, and then decide on unit movement within each group to follow the top level strategy.
\end{Verbatim}
\end{cvbox}
\subsubsection{Example User Request $(n=15, k = 9)$}
\begin{cvbox}
\begin{Verbatim}[breaklines=true, samepage=false]
Goal:
    Surround the target location {Eastfield} from at least three neighboring locations with your units. A neighboring location is considered covered if there are at least 9 units at that location.
Units:
Infantry (Hotel_0) at (Crestview)
Infantry (Hotel_1) at (Meadow)
Infantry (Hotel_2) at (Hillcrest)
Infantry (Hotel_3) at (Hillcrest)
Infantry (Hotel_4) at (Hillcrest)
Infantry (Hotel_5) at (Crestview)
Infantry (Hotel_6) at (Hillcrest)
Infantry (Hotel_7) at (Hillcrest)
Infantry (Hotel_8) at (Hillcrest)
Infantry (Hotel_9) at (Hillcrest)
Infantry (Hotel_10) at (Meadow)
Infantry (Hotel_11) at (Crestview)
Infantry (Hotel_12) at (Hillcrest)
Infantry (Hotel_13) at (Hillcrest)
Infantry (Hotel_14) at (Hillcrest)
Infantry (Lima_0) at (Bayview)
Infantry (Lima_1) at (Creekbend)
Infantry (Lima_2) at (Bayview)
Infantry (Lima_3) at (Townsend)
Infantry (Lima_4) at (Bayview)
Infantry (Lima_5) at (Creekbend)
Infantry (Lima_6) at (Bayview)
Infantry (Lima_7) at (Creekbend)
Infantry (Lima_8) at (Creekbend)
Infantry (Lima_9) at (Creekbend)
Infantry (Lima_10) at (Bayview)
Infantry (Lima_11) at (Creekbend)
Infantry (Lima_12) at (Creekbend)
Infantry (Lima_13) at (Townsend)
Infantry (Lima_14) at (Creekbend)
Infantry (Mike_0) at (Prairie)
Infantry (Mike_1) at (Prairie)
Infantry (Mike_2) at (Prairie)
Infantry (Mike_3) at (Summit)
Infantry (Mike_4) at (Prairie)
Infantry (Mike_5) at (Pineside)
Infantry (Mike_6) at (Summit)
Infantry (Mike_7) at (Prairie)
Infantry (Mike_8) at (Prairie)
Infantry (Mike_9) at (Prairie)
Infantry (Mike_10) at (Prairie)
Infantry (Mike_11) at (Prairie)
Infantry (Mike_12) at (Pineside)
Infantry (Mike_13) at (Summit)
Infantry (Mike_14) at (Summit)

Location Network (location - neighbors):
Eastfield - ['Seabreeze', 'Skyline', 'Moonlight', 'Centerville']
Seabreeze - ['Eastfield', 'Meadow', 'Hillcrest', 'Crestview']
Skyline - ['Eastfield', 'Townsend', 'Bayview', 'Creekbend']
Moonlight - ['Eastfield', 'Summit', 'Pineside', 'Prairie']
Centerville - ['Eastfield', 'Riverbend', 'Lakeside', 'Sunnyside']
Meadow - ['Seabreeze', 'Lakeside', 'Prairie']
Hillcrest - ['Seabreeze', 'Sunnyside', 'Creekbend']
Crestview - ['Seabreeze']
Townsend - ['Skyline', 'Bayview']
Bayview - ['Skyline', 'Pineside', 'Lakeside', 'Riverbend', 'Townsend']
Creekbend - ['Skyline', 'Hillcrest']
Summit - ['Moonlight', 'Pineside']
Pineside - ['Moonlight', 'Summit', 'Bayview']
Prairie - ['Moonlight', 'Meadow']
Riverbend - ['Centerville', 'Bayview']
Lakeside - ['Centerville', 'Bayview', 'Meadow']
Sunnyside - ['Centerville', 'Hillcrest']

\end{Verbatim}
\end{cvbox}


\section{Methods}
\label{app:methods}
\frommak{add clarification that task -> method and effect, effect\_files -> verify and verify\_files}

\subsection{Travel Planner}
\begin{cvbox}

\begin{Verbatim}[breaklines=true, samepage=false]
{
    "method1": { 
        "task": "process user request",
        "subtasks": {
            "subtask1": "take notes on origin, destination, departure date, returning date and other preferences",
            "subtask2": "choose a departing flight",
            "subtask3": "choose a returning flight",
            "subtask4": "choose an accommodation"
        },
        "effect": "answer.txt contains a solution in the correct format that follows the problem specification",
        "effect_files": {
            "file1": "answer.txt",
            "file2": "files/problem_specification.txt"
        }
    },
    "method2": {
        "task": "take notes on origin, destination, departure date, returning date and other preferences",
        "subtasks": {
            "subtask1": "take notes on the problem specification file"
        },
        "effect": "notes contains same details as request and contains origin, destination, departure date and returning date for the trip",
        "effect_files": {
            "file1": "files/notes.txt",
            "file2": "files/request.txt"
        }
    },
    "method3": {
        "task": "choose a departing flight",
        "effect": "notes contains details about a flight number and cost for a flight from origin to destination on departure date consistent with request preferences",
        "subtasks": {
            "subtask1": "understand flights tool"
        },
        "effect_files": {
            "file1": "files/notes.txt",
            "file2": "files/problem_specification.txt",
            "file3": "files/request.txt"
        }
    },
    "method4": {
        "task": "choose a returning flight",
        "effect": "notes contains details about a flight number and cost for a returning flight consistent with request preferences",
        "effect_files": {
            "file1": "files/notes.txt",
            "file2": "files/problem_specification.txt",
            "file3": "files/request.txt"
        }
    },
    "method5": {
        "task": "choose an accommodation",
        "effect": "notes contains details about an accomodation consistent with request preferences", 
        "subtasks": {
            "subtask1": "understand accommodation tool"
        },
        "effect_files": {
            "file1": "files/notes.txt",
            "file2": "files/problem_specification.txt",
            "file3": "files/request.txt"
        }
    },
    "method6": {
        "task": "understand flights tool",
        "effect": "notes contains details about flights tool methods inputs and outputs",
        "effect_files": {
            "file1": "files/notes.txt",
            "file2": "files/tools_specification.txt"
        }
    },
    "method7": {
        "task": "understand accommodation tool",
        "effect": "notes contains details about accommodation tool methods inputs and outputs",
        "effect_files": {
            "file1": "files/notes.txt",
            "file2": "files/tools_specification.txt"
        }
    },
    "method8": {
        "task": "take notes on the problem specification file",
        "effect": "notes contain the information from the problem specification file",
        "effect_files": {
            "file1": "files/problem_specification.txt",
            "file2": "files/notes.txt"
        }
    }
}
\end{Verbatim}
\end{cvbox}

\subsection{Recipe Generator}
\begin{cvbox}
\begin{Verbatim}[breaklines=true, samepage=false]
{
    "method1": {
        "task": "process user request",
        "subtasks": {
            "subtask1": "take notes on user request and problem specification",
            "subtask2": "understand the tools specification"
        },
        "effect": "the answer to the user request can be found in answer.txt",
        "effect_files": {
            "file1": "answer.txt",
            "file2": "files/request.txt",
            "file3": "files/problem_specification.txt"
        }
    },
    "method2": {
        "task": "solve user request",
        "effect": "a valid recipe is chosen",
        "effect_files": {
            "file1": "files/request.txt",
            "file2": "files/problem_specification.txt",
            "file3": "files/notes.txt"
        }
    },
    "method3": {
        "task": "understand the tools specification",
        "effect": "details of provided tools and correct module import path are recorded in notes",
        "effect_files": {
            "file1": "files/tools_specification.txt",
            "file2": "files/notes.txt"
        }
    },
    "method4": {
        "task": "take notes on the problem specification file",
        "effect": "notes contain the information from the problem specification file",
        "effect_files": {
            "file1": "files/problem_specification.txt",
            "file2": "files/notes.txt"
        }
    },
    "method5": {
        "task": "take notes on user request",
        "effect": "notes contain the information from the user request",
        "effect_files": {
            "file1": "files/request.txt",
            "file2": "files/notes.txt"
        }
    },
    "method6": {
        "task": "take notes on user request and problem specification",
        "effect": "notes contain information from the user request and problem specification files",
        "effect_files": {
            "file1": "files/request.txt",
            "file2": "files/problem_specification.txt",
            "file3": "files/notes.txt"
        }
    }
}
\end{Verbatim}
\end{cvbox}

\subsection{Blocks World}
\begin{cvbox}
\begin{Verbatim}[breaklines=true, samepage=false]
{
    
    "method1": { 
        "task": "process user request",
        "subtasks": {
            "subtask1": "take notes on problem specification",
            "subtask2": "take notes on user request",
            "subtask3": "unstack all blocks"
        },
        "effect": "answer.txt contains a solution to the user request in the correct format",
        "effect_files": {
            "file1": "files/problem_specification.txt",
            "file2": "files/request.txt",
            "file3": "answer.txt"
        }
    },
    "method2": {
        "task": "take notes on problem specification",
        "effect": "notes contain the information from the problem specification file",
        "effect_files": {
            "file1": "files/problem_specification.txt",
            "file2": "files/notes.txt"
        }
    },
    "method3": {
        "task": "solve the user request",
        "effect": "notes contain steps to solve the user request",
        "effect_files": {
            "file1": "files/request.txt",
            "file2": "files/problem_specification.txt",
            "file3": "files/notes.txt"
        }
    },
    "method4": {
        "task": "unstack all blocks",
        "effect": "notes contain the intermediate steps to unstack all blocks such that they are all clear and no blocks are being held",
        "effect_files": {
            "file1": "files/request.txt",
            "file2": "files/problem_specification.txt",
            "file3": "files/notes.txt"
        }
    },
    "method5": {
        "task": "take notes on user request",
        "effect": "notes contain a copy of the user request about the initial condition and goal",
        "effect_files": {
            "file1": "files/request.txt",
            "file2": "files/notes.txt"
        }
    }
}
\end{Verbatim}
\end{cvbox}

\subsection{Unit Movement}
\begin{cvbox}
\begin{Verbatim}[breaklines=true, samepage=false]
{
    "method1": {
        "task": "process user request",
        "subtasks": {
            "subtask1": "take notes on problem specification",
            "subtask2": "take notes on user request",
            "subtask3": "group units",
            "subtask4": "move units"
        },
        "effect": "the answer to the user request can be found in answer.txt in the correct format",
        "effect_files": {
            "file1": "answer.txt",
            "file2": "files/request.txt",
            "file3": "files/problem_specification.txt"
        }
    },
    "method2": {
        "task": "move units",
        "subtasks": {
            "subtask1": "move group 1",
            "subtask2": "move group 2",
            "subtask3": "move group 3"
        },
        "effect": "notes contain movement actions to achieve the user request",
        "effect_files": {
            "file1": "files/request.txt",
            "file2": "files/problem_specification.txt",
            "file3": "files/notes.txt"
        }
    },
    "method3": {
        "task": "move group 1",
        "effect": "notes contain moves for units in group 1",
        "effect_files": {
            "file1": "files/notes.txt"
        }
    },
    "method4": {
        "task": "move group 2",
        "effect": "notes contain moves for units in group 2",
        "effect_files": {
            "file1": "files/notes.txt"
        }
    },
    "method5": {
        "task": "move group 3",
        "effect": "notes contain moves for units in group 3",
        "effect_files": {
            "file1": "files/notes.txt"
        }
    },
    "method6": {
        "task": "group units",
        "effect": "notes contain a grouping of units into three distinct groups and a group level strategy for moving each group to achieve the request goal",
        "effect_files": {
            "file1": "files/request.txt",
            "file2": "files/notes.txt"
        }
    },
    "method7": {
        "task": "take notes on user request and problem specification",
        "effect": "notes contain information from the user request and problem specification files",
        "effect_files": {
            "file1": "files/request.txt",
            "file2": "files/problem_specification.txt",
            "file3": "files/notes.txt"
        }
    },
    "method8": {
        "task": "take notes on user request",
        "effect": "notes contain information from the user request file",
        "effect_files": {
            "file1": "files/request.txt",
            "file2": "files/notes.txt"
        }
    },
    "method9": {
        "task": "take notes on problem specification",
        "effect": "notes contain information from the problem specification file",
        "effect_files": {
            "file1": "files/problem_specification.txt",
            "file2": "files/notes.txt"
        }
    }
}
\end{Verbatim}
\end{cvbox}

\section{LLM-generated Task Networks}
\subsection{Blocks World}
\begin{cvbox}
\begin{Verbatim}[breaklines=true, samepage=false]
{
  "method1": {
    "task": "process user request",
    "subtasks": {
      "subtask1": "formulate plan",
      "subtask2": "write final plan to answer.txt"
    },
    "effect": "the answer to the user request can be found in answer.txt in the correct format",
    "effect_files": {
      "file1": "answer.txt",
      "file2": "files/request.txt",
      "file3": "files/problem_specification.txt"
    }
  },
  "method2": {
    "task": "formulate plan",
    "subtasks": {
      "subtask1": "take notes on problem and request",
      "subtask2": "generate plan to clear misplaced blocks",
      "subtask3": "generate plan to build goal stacks"
    },
    "effect": "the full plan to solve the blocksworld problem is stored in notes.txt",
    "effect_files": {
      "file1": "files/notes.txt",
      "file2": "files/request.txt"
    }
  },
  "method3": {
    "task": "take notes on problem and request",
    "subtasks": {
      "subtask1": "take notes on problem specification",
      "subtask2": "take notes on user request"
    },
    "effect": "notes contain the initial state, goal state, and rules of the problem",
    "effect_files": {
      "file1": "files/notes.txt",
      "file2": "files/request.txt",
      "file3": "files/problem_specification.txt"
    }
  },
  "method4": {
    "task": "take notes on problem specification",
    "effect": "notes contain the rules of the blocksworld problem",
    "effect_files": {
      "file1": "files/notes.txt",
      "file2": "files/problem_specification.txt"
    }
  },
  "method5": {
    "task": "take notes on user request",
    "effect": "notes contain the initial and goal block configurations",
    "effect_files": {
      "file1": "files/notes.txt",
      "file2": "files/request.txt"
    }
  },
  "method6": {
    "task": "generate plan to clear misplaced blocks",
    "effect": "notes contains the sequence of unstack and put actions to move all misplaced blocks to the table",
    "effect_files": {
      "file1": "files/notes.txt",
      "file2": "files/request.txt"
    }
  },
  "method7": {
    "task": "generate plan to build goal stacks",
    "effect": "notes contains the sequence of pick and stack actions to build the goal configuration from the blocks on the table",
    "effect_files": {
      "file1": "files/notes.txt",
      "file2": "files/request.txt"
    }
  },
  "method8": {
    "task": "write final plan to answer.txt",
    "effect": "the final, combined plan from notes.txt is written to answer.txt in the correct format",
    "effect_files": {
      "file1": "answer.txt",
      "file2": "files/notes.txt"
    }
  }
}
\end{Verbatim}
\end{cvbox}

\subsection{Unit Movement}
\begin{cvbox}
\begin{Verbatim}[breaklines=true, samepage=false]
{
  "method1": {
    "task": "process user request",
    "effect": "the answer to the user request can be found in answer.txt in the correct format",
    "effect_files": {
      "file1": "answer.txt",
      "file2": "files/request.txt",
      "file3": "files/problem_specification.txt"
    },
    "subtasks": {
      "subtask1": "compose actions",
      "subtask2": "write answer file"
    }
  },
  "method2": {
    "task": "compose actions",
    "effect": "notes.txt contains the JSON list of unit actions for the turn",
    "effect_files": {
      "file1": "files/notes.txt",
      "file2": "files/request.txt",
      "file3": "files/problem_specification.txt"
    },
    "subtasks": {
      "subtask1": "take notes on request and spec",
      "subtask2": "plan strategy",
      "subtask3": "write actions to notes"
    }
  },
  "method3": {
    "task": "take notes on request and spec",
    "effect": "notes.txt contains relevant information extracted from the request and problem specification",
    "effect_files": {
      "file1": "files/notes.txt",
      "file2": "files/request.txt",
      "file3": "files/problem_specification.txt"
    },
    "subtasks": {
      "subtask1": "read request",
      "subtask2": "read problem specification",
      "subtask3": "write notes"
    }
  },
  "method4": {
    "task": "read request",
    "effect": "the contents of files/request.txt have been read and can be used for note taking",
    "effect_files": {
      "file1": "files/request.txt"
    }
  },
  "method5": {
    "task": "read problem specification",
    "effect": "the contents of files/problem_specification.txt have been read and can be used for note taking",
    "effect_files": {
      "file1": "files/problem_specification.txt"
    }
  },
  "method6": {
    "task": "write notes",
    "effect": "files/notes.txt now contains extracted information from the request and specification",
    "effect_files": {
      "file1": "files/notes.txt",
      "file2": "files/request.txt",
      "file3": "files/problem_specification.txt"
    }
  },
  "method7": {
    "task": "plan strategy",
    "effect": "files/notes.txt is updated with a high‑level movement strategy for the infantry units",
    "effect_files": {
      "file1": "files/notes.txt",
      "file2": "files/request.txt",
      "file3": "files/problem_specification.txt"
    }
  },
  "method8": {
    "task": "write actions to notes",
    "effect": "files/notes.txt now contains the final JSON list of unit actions",
    "effect_files": {
      "file1": "files/notes.txt",
      "file2": "files/request.txt",
      "file3": "files/problem_specification.txt"
    }
  },
  "method9": {
    "task": "write answer file",
    "effect": "answer.txt contains the JSON list of unit actions in the required format",
    "effect_files": {
      "file1": "answer.txt",
      "file2": "files/notes.txt",
      "file3": "files/request.txt",
      "file4": "files/problem_specification.txt"
    }
  }
}
\end{Verbatim}
\end{cvbox}

\subsection{Prompt for generating task networks}
The following prompt was used as input to Gemini 2.5 Pro to generate the two LLM-generated task networks above. The empty section labeled task specification is filled with the Problem Spec file for the domain being tested.
\begin{cvbox}
\begin{Verbatim}[breaklines=true, samepage=false]
There is a system that processes commands in order to solve a given task. The task is specified in the problem specification document and the task instance in the request file. An agent is given the following actions to perform:

1. Read: reads [action_arg1]. Line numbers will be displayed.
Read [action_arg1]

2. Write: overwrites text file [action_arg1] with [action_arg2].
Write [action_arg1] [action_arg2]

3. Append: appends the text file [action_arg1] with [action_arg2].
Append [action_arg1] [action_arg2]

4. Verify: Send a request to check if current task has been achieved.
Verify

The agent is allowed to read and write the following file (r/w):
files/notes.txt (r/w): the contents of this file will always be displayed in the notes section down below. 
answer.txt (r/w): a file to hold the final answer to user request

The agent is allowed to access the following read-only files (r):
files/request.txt (r): the user request
files/problem_specification.txt (r): a problem specification document

The agent can only see the contents of the last action it performed (e.g. observations are Markovian), so the agent needs to take frequent notes. The agent will follow a task network with nodes of the form:
{
  "method1": {
      "task": "process user request",
      "subtasks": {
          "subtask1": "[other subtask]",
          "subtask2": "[other subtask]"
      },
      "effect": "the answer to the user request can be found in answer.txt in the correct format",
      "effect_files": {
          "file1": "answer.txt",
          "file2": "files/request.txt",
          "file3": "files/problem_specification.txt"
      }
  },
  ....
  "method7": {
      "task": "take notes on [example1] and [example2]",
      "subtasks": {
          "subtask1": "[other subtask]",
          "subtask2": "[other subtask]"
      },
      "effect": "notes contain information from [example1] and [example2] documents",
      "effect_files": {
          "file1": "files/[example1].txt",
          "file2": "files/[example2].txt",
          "file3": "files/notes.txt"
      }
  }
}
which specify the name of the task, the effect of the task, the files that can be used to confirm the result of the task and possibly a list of subtasks that may need to be completed. For the following problem, please write a task network for nodes of the above format to assist the agent in solving the task.

General instructions:
1. Please respond only with the task network and no additional text. 
2. The agent will need to use the verify action to move onto the next task, so no verification subtasks are needed. 
3. Make sure that the keys for subtasks are numbered backwards in the order that the agent should complete the task in. 
4. All subtasks should have a corresponding task node. The subtasks component is not needed for leaf nodes of the task network.
5. The values in the subtask list "[other subtask]" should correspond to the name of the subtask it is referring to (e.g. "take notes on xyz").
6. The top level task should be called "process user request".
7. Effect files denote the files needed for the verifier to check if a given task node is completed (e.g. if the task is to write an answer, the problem specification and request should be included as effect files so the answer can be checked against it).
8. Make sure each task is numbered task[number].

The task specification:
*** TASK SPEC START ***
{}
*** TASK SPEC END ***
\end{Verbatim}
\end{cvbox}
\section{Verification}
\label{app:verification}

To verify whether an agent LLM has successfully completed a task, we send a request to a verifier LLM to check if the the effect of the task network node has been satisfied. Only the contents of the effect files for a given task node are placed into the context of the verifier LLM. This allows for more consistent evaluation (and alleviates long context issues in using smaller LLMs for verification).
\subsection{Verification Code (success checking in Fig \ref{fig:system-workflow})}
\begin{cvbox}
\begin{Verbatim}[breaklines=true, samepage=false]
with open('verification_prompt.txt') as f:
    lines = f.readlines()
    prompt = "".join(lines)

task_specification = get_current_task_network_node(current_task)
check = task_specification["effect"]

files_to_check = ""
counter = 1
while "file" + str(counter) in task_specification["effect_files"]:
    file_to_check = task_specification["effect_files"]["file" + str(counter)]
    files_to_check += "## Here are the contents of " + file_to_check[:-4] + ": \n"
    with open(files_directory + "/" + file_to_check) as f: 
        lines = f.readlines()
        files_to_check += "".join(lines) + "\n"
    counter += 1
    
prompt = prompt.format(check, files_to_check)
last_output = llm_model.instruct_query(prompt)

success = True if "PASS: TRUE" in last_output else False
if success:    
    predicates.add(stack[-1][0])
    stack = stack[:-1]
    if len(stack) == 0:
        break
    current_task = update_task()
\end{Verbatim}
\end{cvbox}

\subsection{Verification Prompt}
This verification prompt was tuned with meta-prompting on GPT-4.
\begin{cvbox}
\begin{Verbatim}[breaklines=true, samepage=false]
**Here is your task:**  
Your goal is to determine if the provided files satisfy the specified effect. Follow these steps carefully:  

### **Task Specification**  
The effect that needs to be satisfied is:  
{}

### **File Contents**  
Here are the contents of the files provided:  
{}

### **Instructions**  
1. **Understand the Effect:** Carefully read the effect and identify its conditions.  
2. **Analyze the File Contents:** Compare the file contents explicitly against the conditions of the effect.  
    - Break down the effect into specific criteria (e.g., "summarized version" means the main ideas in "request" should be present in "notes").  
    - Check whether these criteria are satisfied in the files.  
3. **Provide a Detailed Analysis:** State which criteria are satisfied and which are not, with reasons based on the file contents.  

### **Your Response Format**  
Begin your response with:  

```
ANALYSIS:  
<Your detailed analysis of whether the files satisfy the effect, with explicit reasoning.>  
```  

End your response with one of the following lines:  

```
PASS: TRUE  
```
*only if all conditions of the effect are satisfied.*  

```
PASS: FALSE  
```
*if any condition of the effect is not satisfied.*  

---
\end{Verbatim}
\end{cvbox}

\end{document}